\documentclass[journal,twoside,web]{ieeecolor}

\usepackage{amssymb}
\usepackage{amsmath}

\usepackage{hyperref}
\usepackage{multirow}%
\usepackage{booktabs}%
\usepackage{booktabs,tabularx}
\usepackage{graphicx} 
\usepackage{multirow} 
\usepackage{subcaption} 
\DeclareCaptionLabelSeparator{ieeeperiod}{.\nobreakspace\nobreakspace}
\usepackage{makecell}
\renewcommand{\thesubsubsection}{\thesubsection\arabic{subsubsection}}

\newcommand{\hide}[1]{}
\usepackage{generic}

\usepackage{cite}
\usepackage{amsmath,amssymb,amsfonts}
\usepackage{algorithmic}
\usepackage{graphicx}
\usepackage{algorithm,algorithmic}
\usepackage{hyperref}
\hypersetup{hidelinks=true}
\usepackage{textcomp}
\begin{document}
\title{Markerless Head Tracking for Accurate and Accessible Neuronavigation}
\author{Ziye~Xie, Oded~Schlesinger, Raj~Kundu, Jessica~Y.~Choi, Pablo~Iturralde, Dennis~A.~Turner, Stefan~M.~Goetz,~\IEEEmembership{Member, IEEE}, Guillermo~Sapiro,~\IEEEmembership{Fellow, IEEE}, Angel~V.~Peterchev,~\IEEEmembership{Fellow, IEEE} and J.~Matias~Di~Martino
\thanks{Research reported in this publication was supported by a Germinator Award from the Duke Institute for Brain Sciences, the National Institute of Mental Health of the National Institutes of Health under Award Number R01MH129733, ONR, Simons Foundation, NSF, and gifts from Apple and Google.
The opinions expressed in this publication are those of the authors and do not necessarily reflect the views of the funding and gifting agencies and companies. \textit{(Ziye Xie, Oded Schlesinger and Raj Kundu are co-first authors; Angel V. Peterchev and J. Matias Di Martino contributed equally to this work.)} \textit{(Corresponding authors: Angel V. Peterchev; J. Matias Di Martino.)}}
\thanks{This work involved human subjects in its research. Approval of all ethical and experimental procedures and protocols was granted by the Duke University Health System Institutional Review Board (Pro00109130).}
\thanks{Ziye Xie, Oded Schlesinger, Stefan M. Goetz, Angel V. Peterchev, and J. Matias Di Martino are with the Department of Electrical and Computer Engineering, Duke University, NC 27708 USA (e-mail: ziye.xie@duke.edu; oded.schlesinger@duke.edu; stefan.goetz@duke.edu; angel.peterchev@duke.edu; matias.di.martino@duke.edu).}
\thanks{Raj Kundu, Jessica Y. Choi, Stefan M. Goetz, and Angel V. Peterchev are with the Department of Psychiatry \& Behavioral Sciences, Duke University, NC 27710 USA (e-mail: raj.kundu@duke.edu; jessica.choi@duke.edu; stefan.goetz@duke.edu; angel.peterchev@duke.edu).}
\thanks{Dennis A. Turner, Stefan M. Goetz, and Angel V. Peterchev are with the Department of Neurosurgery, Duke University, NC 27710 USA (e-mail: dennis.turner@duke.edu; stefan.goetz@duke.edu; angel.peterchev@duke.edu).}
\thanks{Dennis A. Turner and Angel V. Peterchev are with the Department of Biomedical Engineering, Duke University, NC 27708 USA (e-mail: dennis.turner@duke.edu; angel.peterchev@duke.edu).}
\thanks{Pablo Iturralde and J. Matias Di Martino are with the Universidad Católica del Uruguay, 11600 Uruguay (e-mail: pablo.iturralde@ucu.edu.uy; matias.dimartino@ucu.edu.uy).}
\thanks{Raj Kundu is with the Boston University School of Medicine, MA 02118 USA (e-mail: kundu@bu.edu).}
\thanks{Oded Schlesinger and Guillermo Sapiro are with the Department of Electrical and Computer Engineering, Princeton University, NJ 08544 USA (e-mail: odedsc@princeton.edu; guillermos@princeton.edu).}
}
\maketitle

\begin{abstract}
Neuronavigation is widely used in biomedical research and interventions to guide the precise placement of instruments around the head to support procedures such as transcranial magnetic stimulation. Traditional systems, however, rely on subject-mounted markers that require manual registration, may shift during procedures, and can cause discomfort. We introduce and evaluate markerless approaches that replace expensive hardware and physical markers with low-cost visible and infrared light cameras incorporating stereo and depth sensing, combined with algorithmic modeling of the facial geometry. Validation with 50 human subjects yielded a median tracking discrepancy of only 2.32 mm and 2.01$^\circ$ for the best markerless algorithm compared to a conventional marker-based system, which indicates sufficient accuracy for transcranial magnetic stimulation and a substantial improvement over prior markerless results. The study also suggests that integration of the data from the various camera sensors can improve the overall accuracy further. The proposed markerless neuronavigation methods can reduce setup cost and complexity, improve patient comfort, and expand access to neuronavigation in clinical and research settings. 
\end{abstract}

\begin{IEEEkeywords}
Computer vision, Face detection, Head shape modeling, Markerless tracking, Neuronavigation, Pose estimation, Transcranial Magnetic Stimulation.
\end{IEEEkeywords}

\section{Introduction}\label{sec:introduction}
Accurate and robust subject tracking is critical in research and clinical interventions involving brain recording and stimulation. These procedures are typically supported by neuronavigation systems that guide practitioners for targeting specific areas of the brain. The application of neuronavigation, facilitated by the continuous monitoring of a subject's position and orientation, is critical in a variety of procedures. These include precise targeting of brain regions during transcranial magnetic stimulation (TMS)~\cite{goetz2023183}, accurate placement and registration of electroencephalography (EEG) electrodes~\cite{scrivener2022variability}, image-guided surgery~\cite{cleary2010image}, and localization of epileptogenic loci and tumors~\cite{karatacs2004identification, lefaucheur2016value}. Due to the density and complexity of brain tissue, these applications necessitate high accuracy and robustness of neuronavigation to ensure treatment consistency and efficacy.

\begin{figure}[t]
	\centering
	\begin{minipage}[c]{0.49\linewidth}
		\centering
		\includegraphics[height=3.2cm]{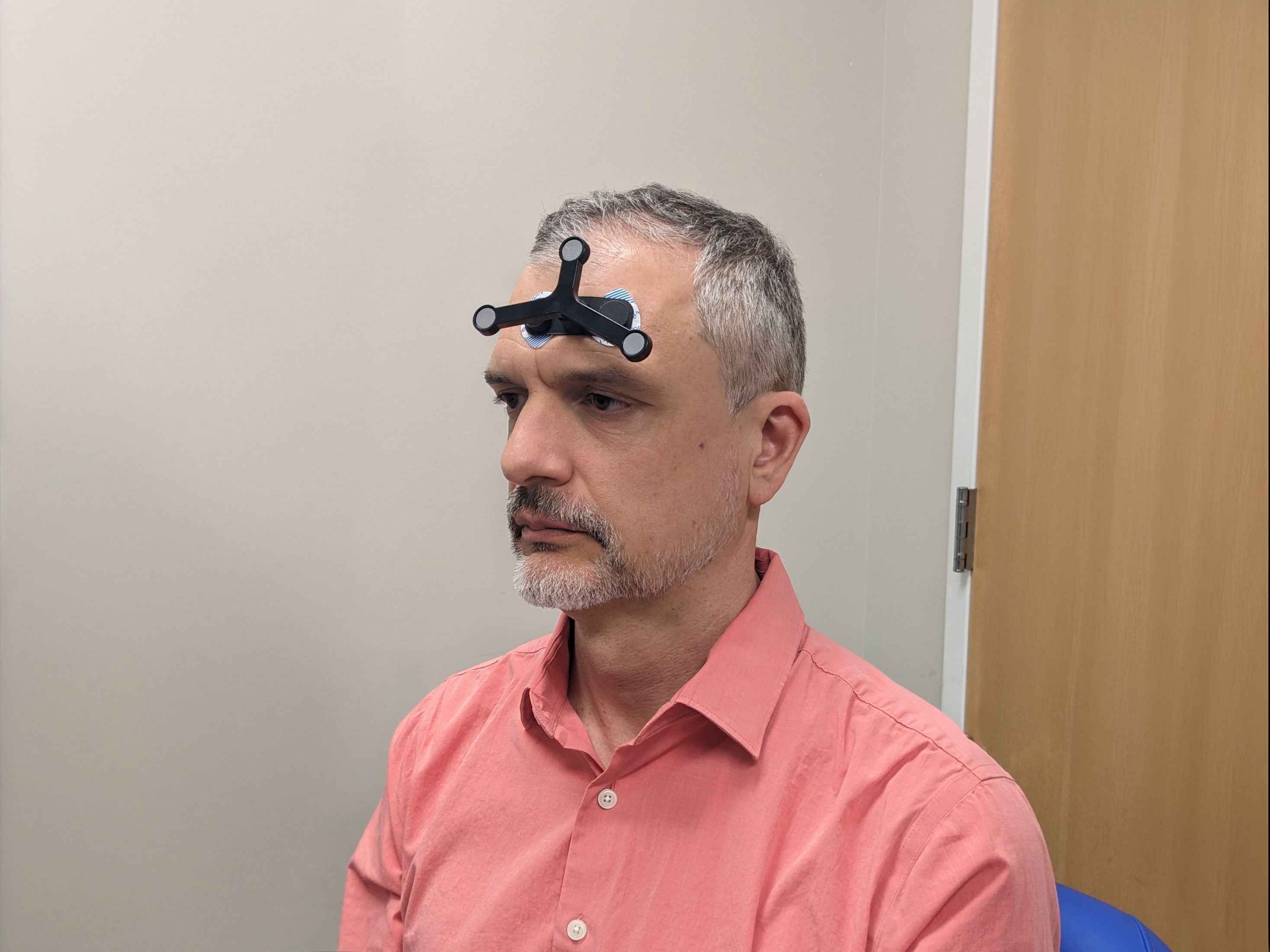}
        \par\vspace{1mm}
        {\footnotesize (a)}
		\label{fig:illustration_a}
	\end{minipage}
	\hfill
	\begin{minipage}[c]{0.49\linewidth}
		\centering
		\includegraphics[height=3.2cm]{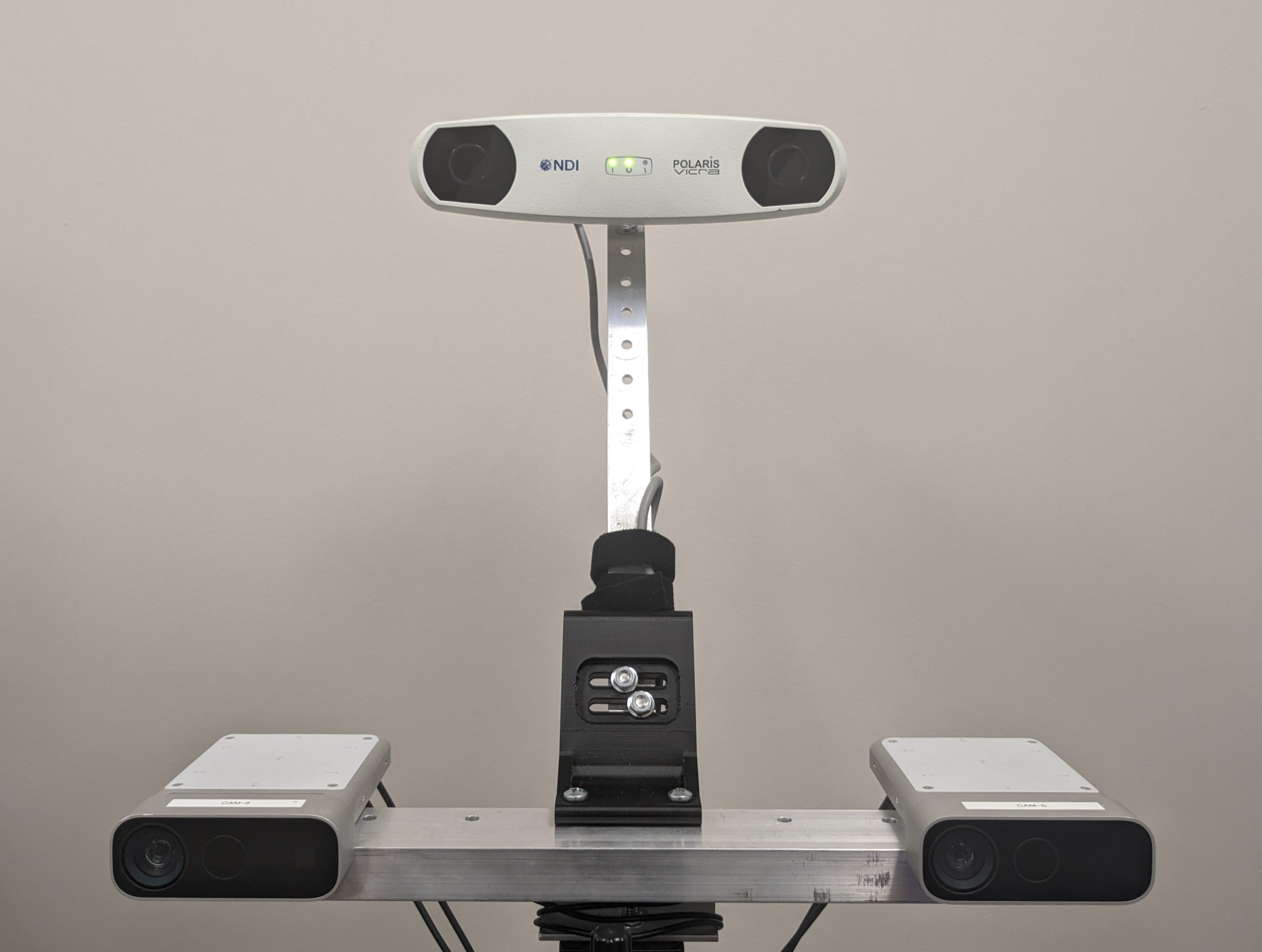}
        \par\vspace{1mm}
        {\footnotesize (b)}
		\label{fig:illustration_b}
	\end{minipage}
    \caption{
    Illustration of conventional marker-based and proposed markerless neuronavigation setups.
    (a) A subject wearing a tracker with retroreflective markers used by commercial neuronavigation systems.
    (b) Experimental apparatus developed in this work, consisting of two Azure Kinect DK cameras (bottom) that capture color and depth data for markerless tracking and, for comparison, a conventional NDI Polaris Vicra stereotaxy camera (top) tracking the retroreflective markers.
    }
    \label{fig:illustration}
\end{figure}

A typical hardware configuration for neuronavigation systems is shown in Fig.~\ref{fig:illustration}. It incorporates a pair of infrared (IR) cameras that localizes passive retroreflective markers in 3D space as the primary tracking device~\cite{neggers_stereotactic_2004, schonfeldt-lecuona_accuracy_2005}. Alternative markers comprise powered electromagnet coils that are tracked with a magnetic field sensor~\cite{wassermann_oxford_2024}. These markers are rigidly attached to the objects of interest, e.g., a patient's head and a TMS coil, and registered manually using a calibration procedure~\cite{ivanov_neuronavigation_2009}.
While the highest tracking accuracy is achieved through rigid fixation to the cranium using stereotactic frames~\cite{maciunas1994application} or skull screws~\cite{holloway2005frameless} to eliminate errors caused by the skin shifting several millimeters relative to the skull~\cite{maurer1997registration}, such methods are limited to invasive interventions.
Consequently, marker attachment to human subjects is commonly implemented with goggles, elastic headbands, or adhesive pads applied to the skin.
The registration process commonly entails manual identification of specific anatomical landmarks and subsequent establishment of the spatial relationship between these landmarks and the markers using a pointing device~\cite{wassermann_oxford_2024}. Following registration, the neuronavigation system can infer the position and orientation of the objects of interest by tracking the affixed physical markers.

Factors limiting the accuracy and robustness of neuronavigation-guided procedures stem from their dependence on physical markers and the manual registration protocols, as well as the lack of personalized head models and head 3D scanning. The premise of precise localization of a tracked object is the rigid attachment of the markers to the subject; however, markers are susceptible to displacement~\cite{nieminen2022accuracy, mitsui2011skin} and may exhibit drift throughout a procedure~\cite{wassermann_oxford_2024, goetz_accuracy_2019}. In addition, the use of adhesive pads might introduce patient discomfort or induce allergic reactions~\cite{barton_medical_2024, thornton2021contact}, while alternatives such as goggles and elastic bands may not be able to maintain rigid attachment to the subject. Beyond the limitations introduced by physical markers, the manual registration of anatomical landmarks is a time-consuming and error-prone process that introduces human error~\cite{houck2020comparison} and is susceptible to displacement of soft tissues when pressed by a pointer~\cite{wassermann_oxford_2024}. These discrepancies may lead to significant reductions in tracking performance, and the reliance of neuronavigation systems on specialized markers and tracking sensors significantly increases cost and limits their adoption.

To address these limitations, we employ computer-vision-based techniques for neuronavigation systems by leveraging the human face as a natural tracker to develop accurate and comfortable head tracking methods. We developed a multi-modal consumer-grade hardware setup (described in detail in Appendix~\ref{supp:subsec:experimental_setup}) to examine different methods for tracking the head movements of human subjects and compare them to data captured by a conventional neuronavigation system. Our methods employ recent advances in face detection, facial landmark localization, and 3D head statistical modeling, and establish the foundation for an accurate and robust markerless neuronavigation system. In addition, our work has the potential of reducing human error and setup complexity while lowering the financial barriers associated with neuronavigation systems. 

The contributions of this work include
(i) three computer-vision-based methods for subject head tracking in neuronavigation utilizing one or two consumer-grade RGB-D cameras,
(ii) the novel utilization of individualized, statistical prior-based head models for head tracking in neuronavigation,
(iii) a controlled and direct systematic evaluation of the proposed computer-vision-based methods in comparison with a conventional neuronavigation solution, and
(iv) an extensive experimental assessment of the proposed methods on a cohort of $50$ human subjects in a treatment-suite environment.

\section{Related Work}\label{sec:related_work}
A growing body of research has explored markerless tracking approaches as a solution to the limitations of conventional marker-based neuronavigation systems. Several studies have focused on replacing physical markers with anatomical registration techniques using facial geometry. Chiurillo et al.~\cite{chiurillo2023high} demonstrated a markerless optical neuronavigation system based on Computed Tomography (CT) scans and 3D images registration of facial anatomy, while Kögl et al.~\cite{kogl2023tool} proposed a markerless monocular-camera approach for annotating fiducial landmarks to facilitate subject registration. Other works have used augmented reality platforms to integrate facial landmark-based registration with head-mounted self-localization systems~\cite{gsaxner2019markerless}, and Sathyanarayana et al.~\cite{sathyanarayana2020comparison} compared marker-based against markerless tracking in mixed-reality neuronavigation, which supported the clinical relevance of markerless methods.
Beyond these approaches, other lines of work have explored computer-vision frameworks that exploit anatomical structures and features for markerless tracking. Manni et al.~\cite{manni2020towards} proposed a method for optical-image pre-processing in spinal surgery by detecting anatomical landmarks and matching local invariant features, whereas Liu et al.~\cite{liu2017new} introduced an automatic image-to-patient registration system for neurosurgical interventions for efficient alignment of preoperative imaging and real-time patient anatomy. Zeng et al.~\cite{zeng2023markerless} demonstrated markerless event-by-event motion correction in brain PET imaging utilizing stereo IR cameras to capture 3D facial surfaces. Huang et al.~\cite{huang2024intercap} introduced a joint 3D tracking framework that leverages multi-view RGB-D input to capture detailed poses of humans and objects in contact-rich scenarios. Similarly, Cuevas et al.~\cite{cuevas2025mamma} proposed a multi-view markerless motion capture pipeline for modeling two-person interaction dynamics via parametric body shape estimation.
These methods demonstrate the potential of integrating computer-vision techniques to replace traditional marker-based registration and tracking in the biomedical context.

Recent studies have demonstrated the feasibility of employing the Azure Kinect DK in a range of clinical applications, including human body motion analysis~\cite{brambilla2023azure}, subject positioning~\cite{bertram2023accuracy}, and joint motion and angle tracking~\cite{villa2025benchmarking}. Notably, the Azure Kinect DK is available at a relatively low cost of \$399 (2025 retail price), positioning it as a scalable alternative to specialized tracking hardware. However, its application to head tracking for neuronavigation, particularly in light of the associated high-precision requirements, remains an active area of research.

To the best of our knowledge, the most relevant previous work on computer-vision-based markerless neuronavigation is the MarLe study~\cite{matsuda2023marle}, which proposed a monocular RGB-based head pose tracking method.
They assessed the stability of pose estimation using a printed face photo, and subsequently demonstrated an integrated neuronavigation workflow for repositioning a TMS coil. 
Although this study showed the potential for markerless neuronavigation, its empirical validation was limited and included a single subject. 
Our study advances significantly beyond this initial effort. 
First, our work evaluates additional modalities such as stereo and depth video data, and they appear key to accurate tracking according to our results. 
Second, our methods incorporated facial statistical priors, including a generic face model and an individualized head model.
Third, we conduct a comprehensive evaluation of our methods in a cohort of 50 subjects and report direct comparison results against conventional neuronavigation hardware.

\section{Methods}\label{sec:methods}
\subsection{Study Participants}\label{subsec:study_participants}  

Fifty healthy participants completed the study (27 female, 23 male, 20--76 years old, with a mean of 28.6). An effort was made to recruit participants of various racial and ethnic backgrounds to capture a diverse set of facial features for robust development and evaluation of the tracking algorithms. The demographic composition of the subjects is detailed in Table~\ref{table:composition}.
\begin{table*}[t]
\caption{Demographic Composition of Study Subjects}
\label{table:composition}
\centering
\resizebox{1\textwidth}{!}{
\begin{tabular}{|cc|ccc|ccccc|cc|}
\hline
\multicolumn{2}{|c|}{Gender (\# subj)} &
\multicolumn{3}{c|}{Age (years)} &
\multicolumn{5}{c|}{Race (\%)} &
\multicolumn{2}{c|}{Ethnicity (\%)} \\ \hline
   \multirow{3}{*}{Female} 
 & \multirow{3}{*}{Male} 
 & \multirow{3}{*}{Min}& \multirow{3}{*}{Mean} 
 & \multirow{3}{*}{Max}& \multirow{3}{*}{White} 
 & \multirow{3}{*}{\parbox{4.5em}{\centering Black or African American}}
 & \multirow{3}{*}{Asian} 
 & \multirow{3}{*}{\parbox{4.5em}{\centering More than one race}} 
 & \multirow{3}{*}{\parbox{6em}{\centering American Indian or Alaska Native}} 
 & \multirow{3}{*}{\parbox{3.5em}{\centering Hispanic or Latino}} 
 & \multirow{3}{*}{\parbox{4em}{\centering Not Hispanic or Latino}} \\
 & & & & & & & & & & & \\ 
 & & & & & & & & & & & \\ \hline
 \rule{0pt}{2ex}27 & 23 & 20 & 28.6 & 76 & 38 & 30 & 26 & 4 & 2 & 26 & 74 \\ \hline
\end{tabular}%
}
\par\vspace{2pt}
\begin{minipage}{\dimexpr\linewidth-2\tabcolsep-2\arrayrulewidth\relax}
\footnotesize
\raggedright
Participants were volunteers from the general population.
\end{minipage}
\end{table*}

\subsection{Experimental Setup}\label{subsec:experimental_setup}  
Participants were instructed to sit in front of the data collection apparatus shown in Fig.~\ref{fig:illustration}, which consists of an NDI Polaris Vicra stereotaxy camera (NDI) on top of a pair of rigidly attached and calibrated Microsoft Azure Kinect DK (Azure) RGB-D camera devices. The NDI stereotaxy camera is a conventional tracking device for neuronavigation systems. It tracks a retroreflective Brainsight Subject Tracker (marker) mounted on the subject. The pair of Azure devices operates in wired synchronization mode and stays independent of the NDI camera. Collectively, time-synchronized and spatially co-registered RGB and depth data streams were recorded by each Azure device. Appendix~\ref{supp:subsec:experimental_setup} provides further configuration details. 

Two recordings were acquired for each participant, who were instructed to maintain a neutral facial expression while performing a series of controlled head movements.
In the first recording, subjects were instructed to perform two sets of clockwise and counterclockwise head roll movements. This recording facilitated the comprehensive facial data acquisition by the Azure system and is referred to as the ``face scan recording''.
After the face scan recording, a retroreflective subject tracker was affixed to the forehead of subjects with adhesive EEG snap electrodes as shown in Fig.~\ref{fig:illustration}(a), following standard practice for TMS neuronavigation. Subsequently, subjects were instructed to replicate eight different head movements by following along a demonstration video. These movements include translation (sway, surge, and heave) and rotation (roll, pitch, and yaw), as well as combined pitch and yaw rotations for clockwise and counterclockwise head roll movements. This second recording was used for tracking accuracy analysis, during which concurrent data acquisition was performed using both the NDI and Azure systems.

Data frames in which the operator inadvertently obscured the subject from the data collection apparatus, or when the retroreflective markers detached from the forehead of the subject due to failure of the adhesive electrode pads, were excluded from further analysis. These issues were observed in data from six (out of 50) subjects, and the average frame exclusion rate in those six videos was $22.4\%$ ($9,298$ out of $41,419$ frames). Importantly, these six subjects remained in the study analysis, and only the specific affected frames were discarded. Overall, the total number of frames included in the comparison between methods was $336,755$.

\subsection{Tracking Methods}\label{subsec:tracking_methods}
\begin{figure}[t]
    \centering
    \includegraphics[width=\linewidth]{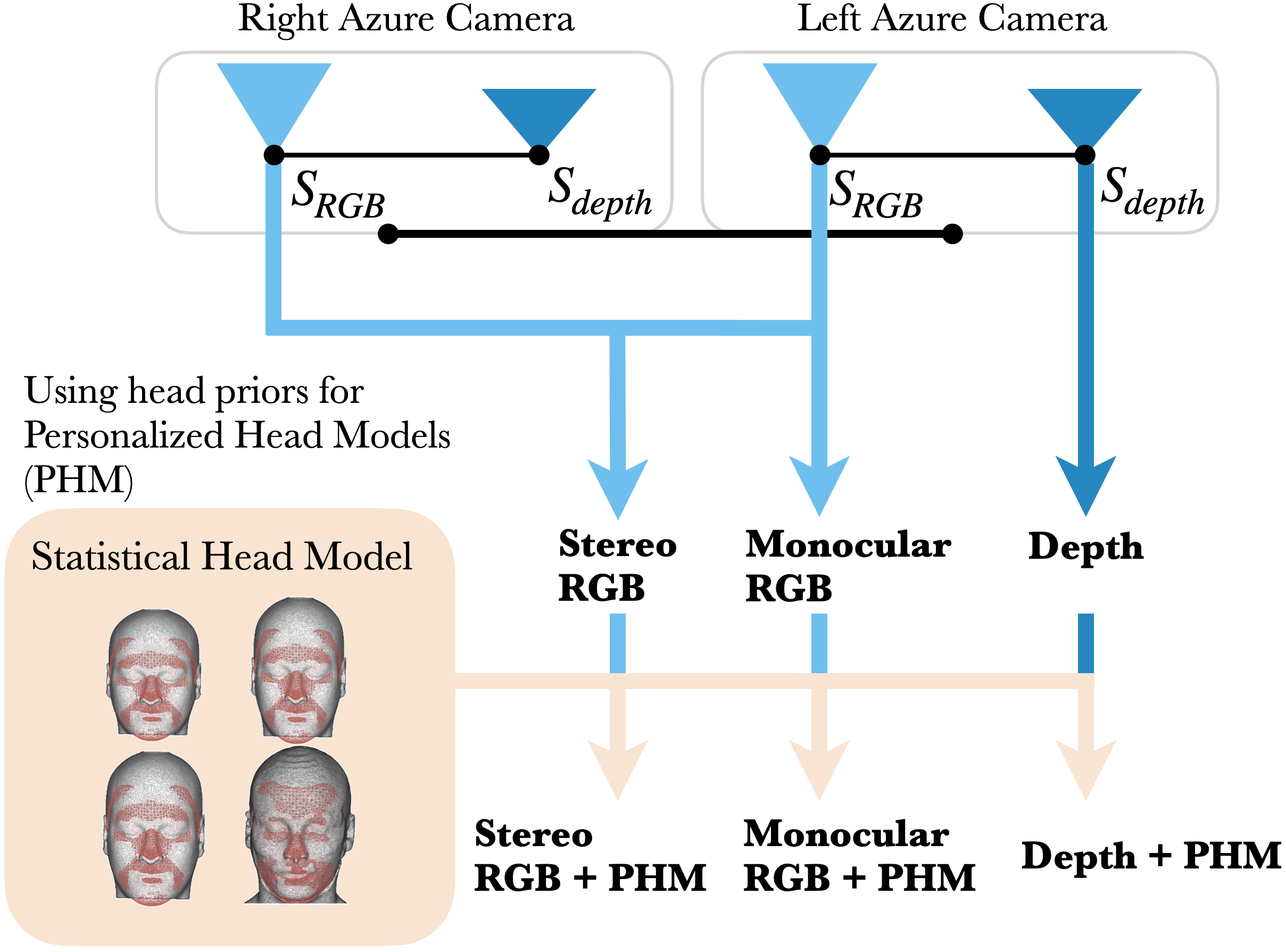}
    \caption{The markerless tracking system comprises a pair of Azure Kinect DK devices.
    The devices are synchronized and spatially calibrated (thick horizontal black line).
    Each device contains an RGB and a depth camera (labeled as $S_{RGB}$ and $S_{depth}$, respectively), operating simultaneously (thin horizontal black lines).
    We implemented and compared three neuronavigation markerless tracking strategies: Monocular RGB (Section~\ref{subsubsec:rgb_mono_method}), Stereo RGB (Section~\ref{subsubsec:rgb_stereo_method}), and Depth (Section~\ref{subsubsec:depth_tracking}). We compared these techniques with and without the use of statistical head priors (Section~\ref{subsubsec:head_modeling}), leading to a final set of six tracking alternatives.}
    \label{fig:tracking_methods}
\end{figure}
We tracked the pose of the subject's head using a combination of the RGB and depth data from a pair of Azure devices, as shown in Fig.~\ref{fig:tracking_methods}. Each Azure device acquired two data modalities: color images from the RGB camera and point clouds from the depth sensor. We exploited this multimodal information to compare three tracking methodologies: tracking through a single RGB camera (Monocular RGB), a stereo RGB camera pair (Stereo RGB), and a depth sensor (Depth). Moreover, we explored the use of statistical head priors to regularize the problem of head shape estimation, leading to a final set of six tracking alternatives (each of the three methodologies defined above with and without the use of statistical head priors).

\subsubsection{Notation}\label{subsec:notation}
In this study, we use the following notation conventions: camera and world points of interest are represented in homogeneous coordinates and denoted using bold lowercase variables, e.g., $\mathbf{w}=(u,v,1)^T$ represents the $(u,v)$ pixel coordinates in a given camera sensor, and $\mathbf{x} = (x, y, z, 1)^T$ represents the $(x,y,z)$ world coordinates in a given reference frame. When the reference frame is not obvious by context, we will introduce a subindex to specify it. Ordered sets of points, such as the set of all landmarks corresponding to a face, are denoted using uppercase variables in both 2D and 3D, e.g., $W = \left[  \mathbf{w}_1, \mathbf{w}_2, \ldots, \mathbf{w}_n \right]$ and $X = \left[ \mathbf{x}_1, \mathbf{x}_2, \ldots, \mathbf{x}_n \right]$. Unordered collections of points, such as point clouds, are denoted using calligraphic notation, e.g., $\mathcal{X} = \lbrace \mathbf{x}_1, \mathbf{x}_2, \ldots, \mathbf{x}_n \rbrace$.
\begin{figure}[t]
    \centering
    \begin{tabular}{@{}c@{\hspace{0.0005\linewidth}}c@{}}
        \includegraphics[scale=0.049]{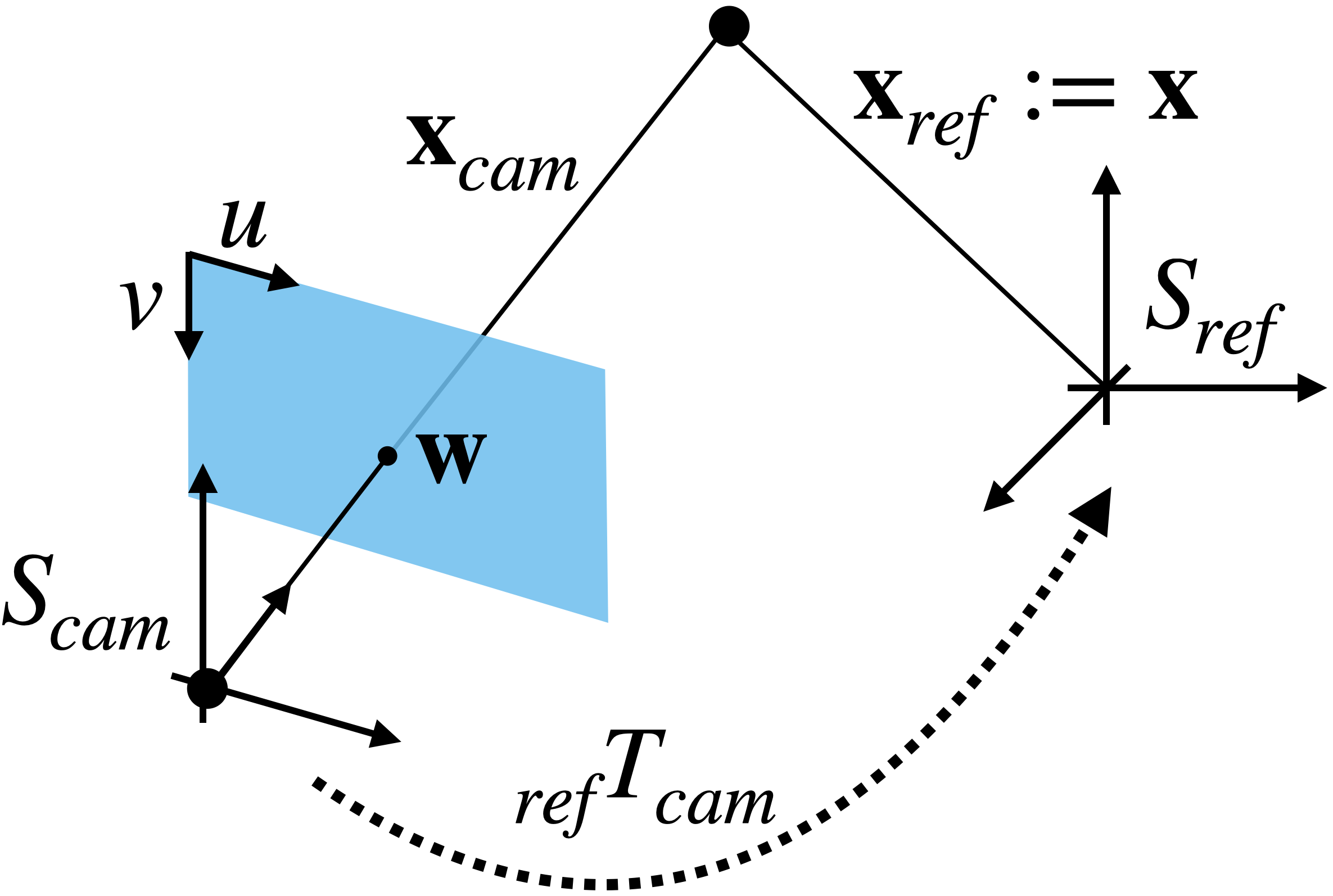} &
        \includegraphics[scale=0.049]{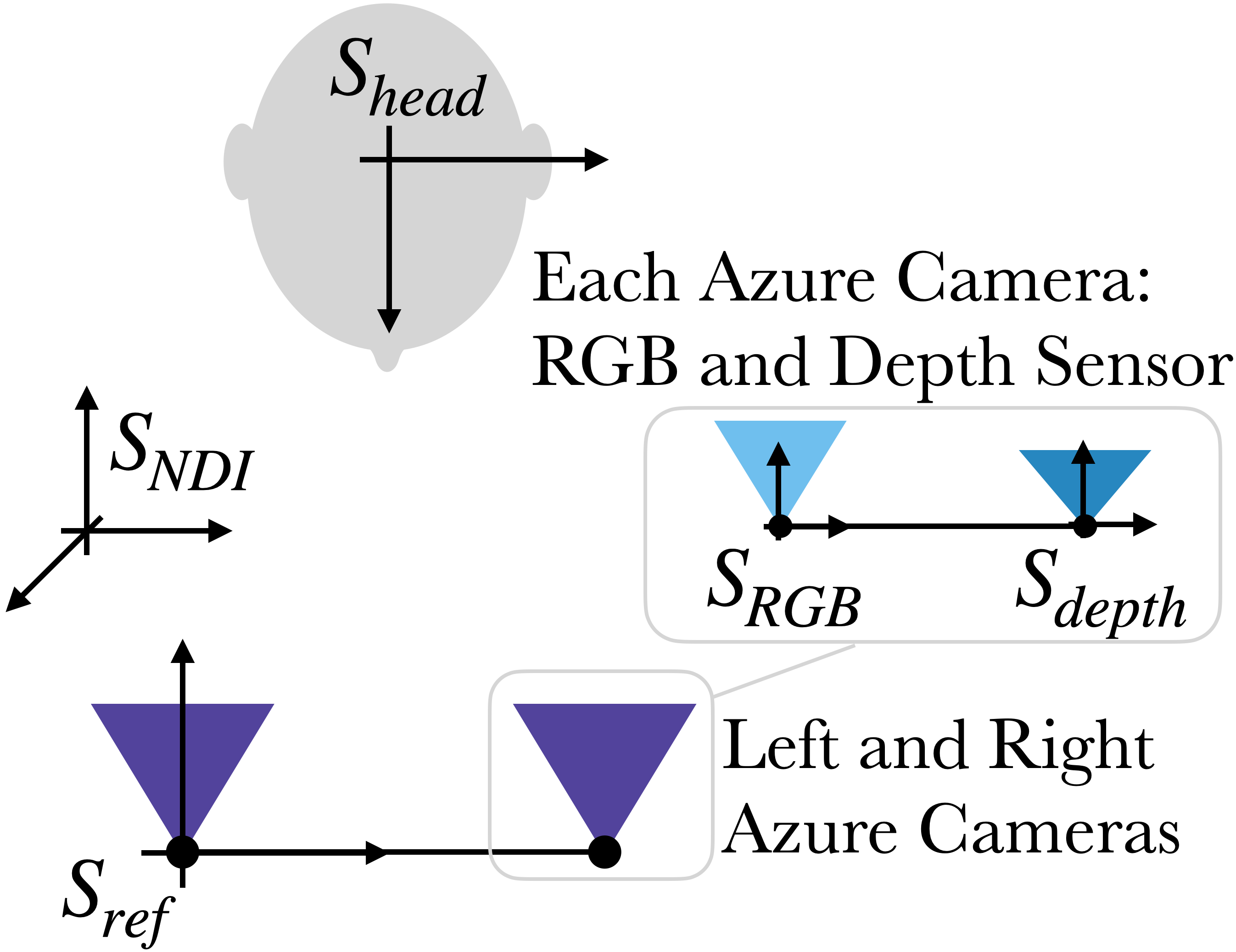} \\
        {\footnotesize (a)} & {\footnotesize (b)}
    \end{tabular}
    \caption{Notation style and reference systems. (a) We adopt a pinhole camera model to represent the geometry of the captured data in the camera frames ($S_{cam}$) and reference coordinate frames ($S_{ref}$). Subindices will be used to denote the reference frame in which points, sets of points, and point clouds are expressed. 3D transformations between frames are represented as ${}_{destination}T_{source}:=\ S_{source}\rightarrow S_{destination}$. (b) Each Azure device has two sensors and associated coordinate frames (RGB and depth). The head keypoint will be represented in a coordinate frame fixed to the head ($S_{head}$). External data collected for validation, e.g., from a conventional NDI system, has its own reference frame ($S_{NDI}$).}
    \label{fig:notation}
\end{figure}

Five distinct reference systems are considered. The first three are used in our tracking algorithms: one corresponding to the left Azure sensor ($S_{left}$), one corresponding to the right Azure sensor ($S_{right}$), and one corresponding to the tracked subject's head ($S_{head}$). We define the RGB frame as the primary reference frame for each Azure device, mapping depth data to the RGB frame when necessary. We define $S_{left}$ as the worldview reference frame $S_{ref}$, to ensure consistency across all experiments. The other two reference systems are employed by the NDI system, which we use for comparison: the NDI stereotaxy camera ($S_{NDI}$) and the retroreflective marker set that it tracks ($S_{marker}$). We also use subscripts to denote the reference system in which points, sets of points, and point clouds are expressed. For example, $\mathbf{X}_{cam}$ refers to a set of points described in the reference system $S_{cam}$, and when no reference system is specified, we assume the reference system $S_{ref}$. Transformations between reference frames are noted as ${}_{destination}T_{source}:=\ S_{source}\rightarrow S_{destination}$, and are represented by a $4 \times 4$ transformation $T = [R | \mathbf{t}]$ in homogeneous coordinates. Illustration of the notation style and reference systems is shown in Fig.~\ref{fig:notation}.


\subsubsection{Face Detection and Landmark Localization}\label{subsubsec:face_alignment}
\begin{figure}[t]
    \centering
    \begin{tabular}{@{}c@{\hspace{0.008\linewidth}}c@{\hspace{0.008\linewidth}}c@{}}
        \includegraphics[height=3.12cm]{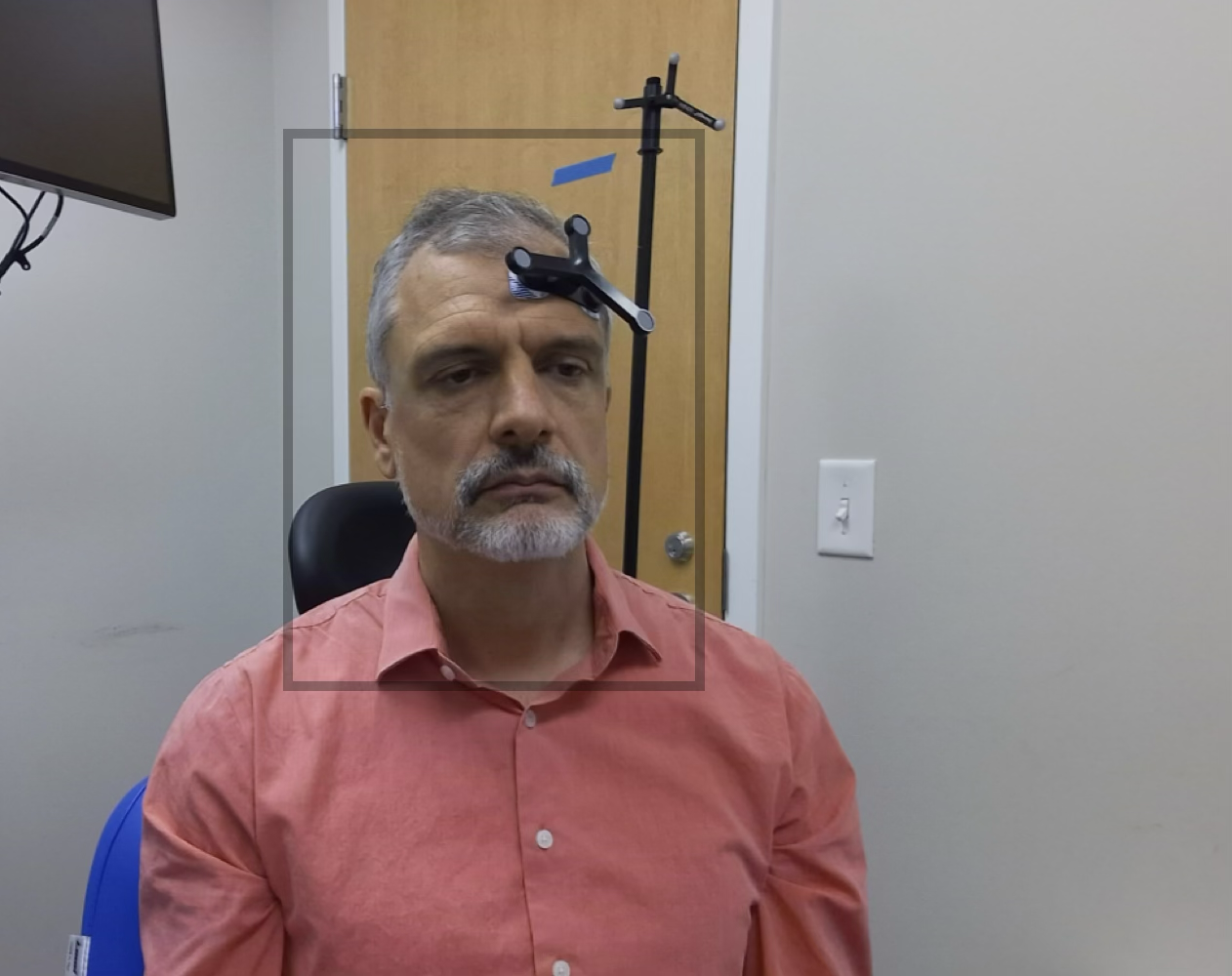} &
        \includegraphics[height=3.12cm]{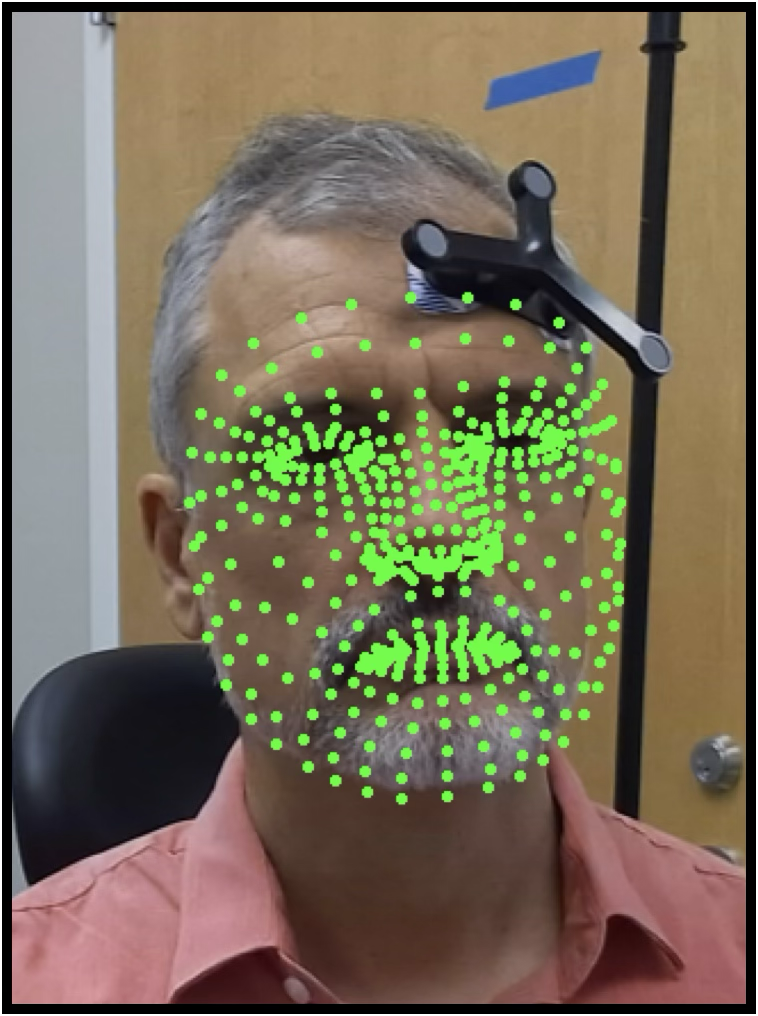} &
        \includegraphics[height=3.12cm]{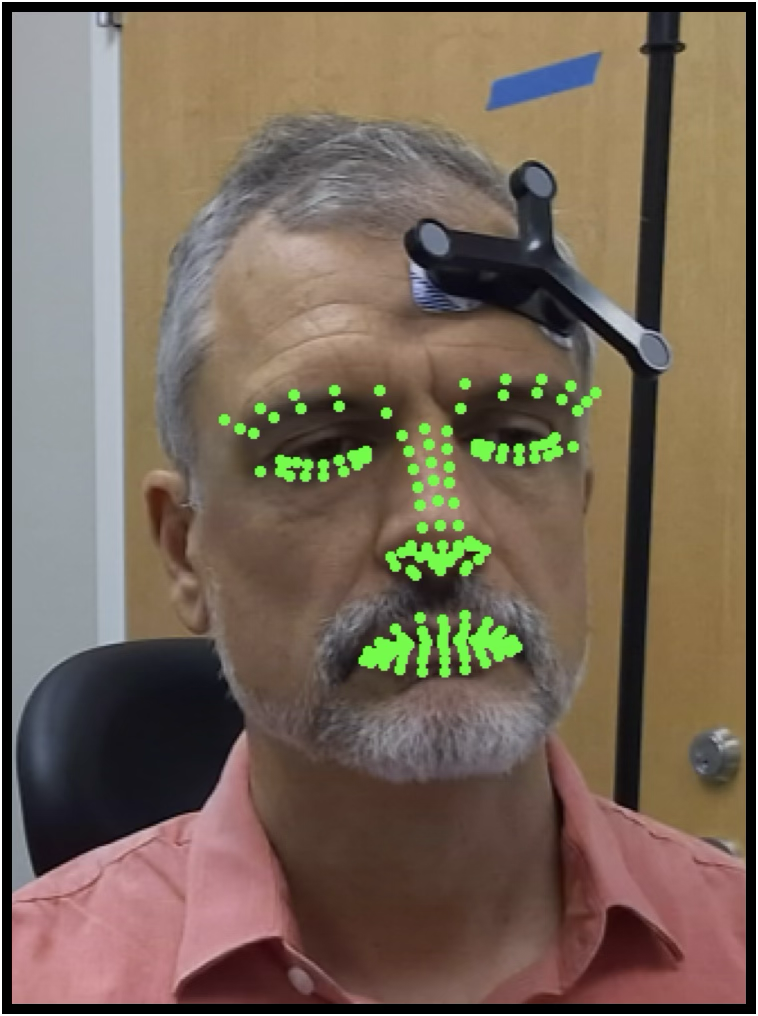} \\
        {\footnotesize (a)} & {\footnotesize (b)} & {\footnotesize (c)}
    \end{tabular}
    \caption{Face detection and landmark localization. (a) Sample RGB data frame from the reference Azure camera. (b) Zoomed-in view of (a) showing the set of detected MediaPipe facial landmarks. (c) Zoomed-in view of (a) showing the subset of landmarks used for tracking.}
    \label{fig:landmarks_rgb}
\end{figure}

We preprocess our data by performing face detection and facial landmark localization using the MediaPipe~\cite{kartynnik_real-time_2019} framework, as shown in Fig.~\ref{fig:landmarks_rgb}. MediaPipe localizes 468 facial landmarks in an RGB image and supplies a 3D canonical face model. Facial landmarks associated with areas of the face that are visually salient and have strong semantic definition (e.g., the corner of the eyes) tend to be more reliably localized, as they suffer less from annotation ambiguity compared to weakly defined landmarks along contours~\cite{liu2019semantic}. We empirically identified and selected a subset of these salient facial landmarks and will refer to this subset of landmarks as the MediaPipe-Subset $W$ and $X^{\text{template}}$, where they respectively denote the set of detected 2D facial landmark pixel $\mathbf{w_i}$ and the 3D MediaPipe canonical face model landmark $\mathbf{x_i}$ coordinates. 
Details on the selection of this optimal set and ablation studies can be found in Appendix~\ref{supp:subsec:rgb_ablation_study}. 
\\

\subsubsection{Monocular RGB-Based Tracking}\label{subsubsec:rgb_mono_method}
Perspective-n-Point (PnP) algorithms are extensively used methods for monocular object pose estimation~\cite{madsen2004methods, dementhon_model-based_1995, haralick_review_1994, lepetit_epnp_2009}. These algorithms take a set of 2D image points and their corresponding locations in the object of interest's 3D model as input, and estimate the pose of the object by minimizing the 2D reprojection discrepancy between these 2D and 3D correspondences. 
In the context of head pose estimation, given an RGB image, these methods require the detection of a set of 2D facial landmarks, e.g., $W$, and the supply of a 3D face model, e.g., $X^{\text{template}}$.\\
For each RGB camera, we calculated the projection matrix $P$ through standard camera calibration procedures (whose details could be found at Appendix~\ref{supp:subsubsec:camera_calibration}). This allowed us to project 3D coordinate points into the camera pixel space, i.e., $z \mathbf{w} = P \mathbf{x}$, where $z$ is a scalar factor representing point depth along the optical axis of the camera frame $S_{left}$. The canonical formulation assumes 3D vectors are described in the camera frame; the reprojection equation can be generalized to any head model frame $S_{head}$ by including a transformation between the head and camera frame, i.e., $z \mathbf{w} = P\,{}_{left}T_{head}\,\mathbf{x}_{head}$.

The monocular head pose estimation problem can then be formulated as finding the transformation matrix ${}_{ref}T_{head}$ defined by the optimization problem:
\begin{equation}
{}_{ref}T_{head} = \mbox{argmin}_{T\in SE(3)} \| W - \sigma \left( P T X^{\text{template}}\right) \|^2, 
\label{eq:monocular_view}
\end{equation}
where the operator $\sigma ()$ refers to the column-wise normalization of a matrix using its last row to project numerical solutions into its homogeneous coordinates.

\subsubsection{Stereo RGB-Based Tracking}\label{subsubsec:rgb_stereo_method}
The spatial location of a point can be determined through the principle of stereo triangulation~\cite{Hartley2004}, which infers 3D information via 2D pixel point correspondence. We applied this principle to reconstruct the 3D coordinates of a set of distinct facial landmarks identified in the pair of RGB cameras available in each Azure device. We modeled the head as a rigid body; hence, we can estimate the head pose by finding the optimal transformation that aligns the estimated 3D landmark locations with a reference 3D face model. 

Initially, the 2D pixel coordinates of a set of MediaPipe landmarks were estimated for both the left and the right RGB images, as described in Section~\ref{subsubsec:rgb_mono_method}. We denote $\mathbf{w_{left}}$ and $\mathbf{w_{right}}$ the 2D pixel coordinates detected in the left and right RGB cameras for one landmark. The RGB cameras were calibrated (detailed at Appendix~\ref{supp:subsubsec:camera_calibration}) to obtain the projection matrices $P_{left}$ and $P_{right}$.
The location of each landmark in the 3D world $\mathbf{x}$ was then obtained by solving:
\begin{equation}
    \mathbf{w_{left}} \times P_{left}\,\mathbf{x} = 0, \quad \mathbf{w_{right}} \times P_{right}\,\mathbf{x} = 0.
    \label{eq:triangulation}
\end{equation}
Equation~\eqref{eq:triangulation} can be rearranged into a linear system of equations whose solution is obtained through Singular Value Decomposition (SVD)~\cite{Hartley2004}. This procedure was repeated for each landmark to obtain a set of 3D points expressed in the left camera reference frame, denoted as $X$.

With the 3D landmarks reconstructed, the head pose ${}_{ref}T_{head}$ can be estimated by finding the optimal transformation that aligns a template 3D face model to these landmarks:
\begin{equation}
{}_{ref}T_{head} = \mbox{argmin}_{T\in SE(3)} \| X - \left( T\, X^{\text{template}} \right) \|^2.
\label{eq:stereo_face_registration}
\end{equation}

\subsubsection{Depth-Based Tracking}\label{subsubsec:depth_tracking}
We utilized depth data from the reference Azure device to estimate the head pose. For each data frame, a point cloud $\mathcal{X}$ from the depth sensor is used to estimate the transformation that aligns a reference facial geometry $\mathcal{X}^{\text{template}}$ to the captured point cloud corresponding to that data frame. Unlike previous methods that rely on sparse sets of landmarks, this method leverages the full dense point cloud data captured by the depth sensor.

\begin{figure}[t]
    \centering
    \begin{tabular}{@{}c@{\hspace{0.008\linewidth}}c@{\hspace{0.008\linewidth}}c@{}}
        \includegraphics[height=3.12cm]{Figures/MediaPipe_landmarks_edited2_a.png} &
        \includegraphics[height=3.12cm]{Figures/MediaPipe_landmarks_edited2_b.png} &
        \includegraphics[height=3.12cm]{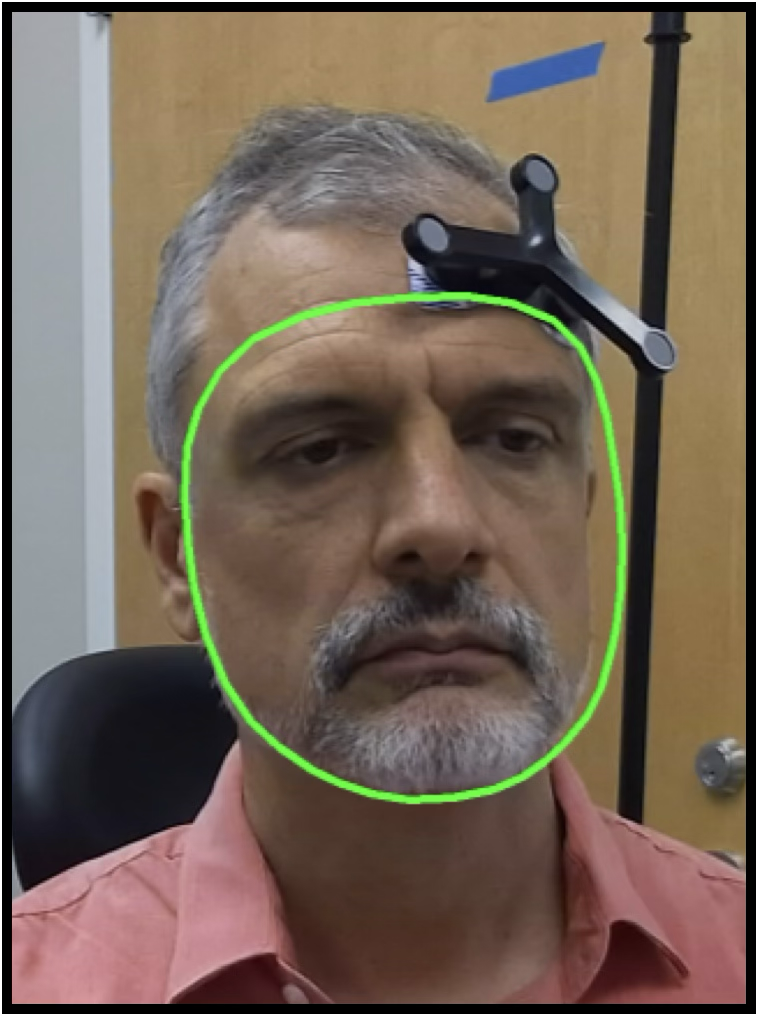} \\
        {\footnotesize (a)} & {\footnotesize (b)} & {\footnotesize (c)}
    \end{tabular}
    \vspace{1mm}
    \begin{tabular}{@{}c@{\hspace{0.008\linewidth}}c@{\hspace{0.008\linewidth}}c@{}}
        \includegraphics[scale=0.08]{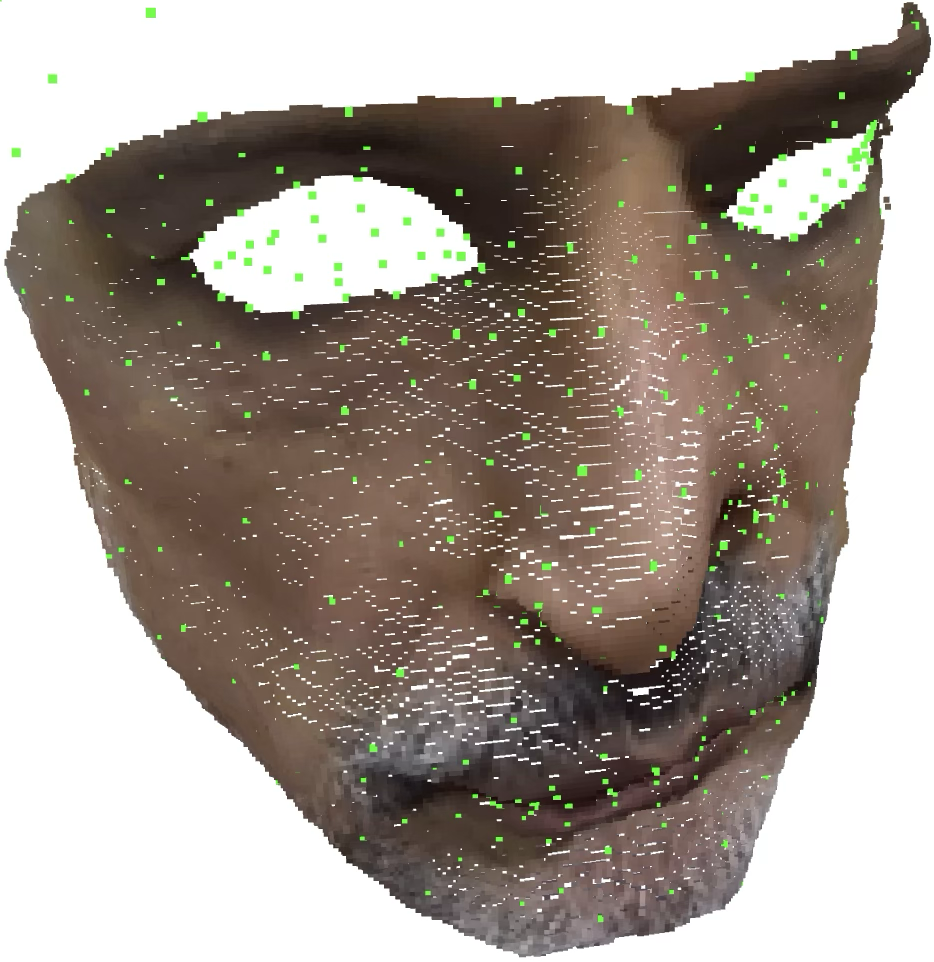} &
        \includegraphics[scale=0.085]{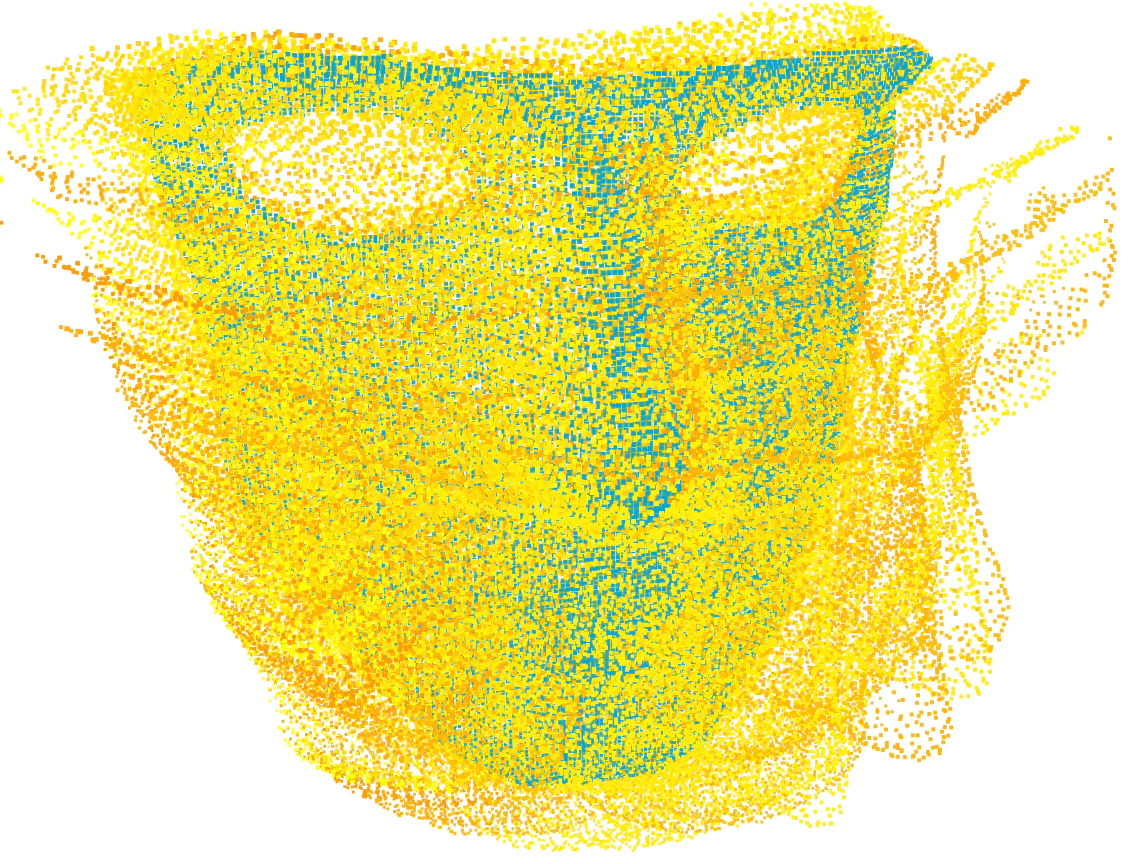} &
        \includegraphics[scale=0.085]{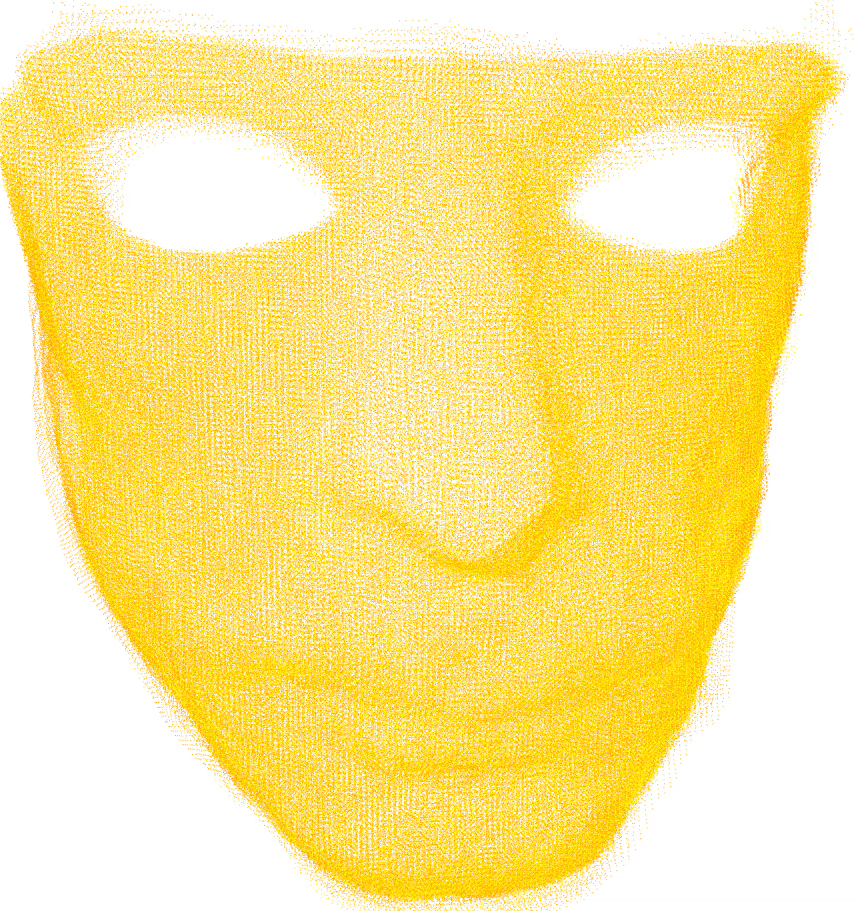} \\
        {\footnotesize (d)} & {\footnotesize (e)} & {\footnotesize (f)}
    \end{tabular}
    \caption{Creation of the 3D facial template for point cloud tracking. (a) Sample RGB data frame. (b) MediaPipe landmarks detected in the RGB image. (c) Convex hull of the landmarks used to define the facial region of interest. (d) Color information is mapped into the 3D reference frame to select the point cloud region of interest. (e) Point clouds obtained during a face scan recording are aligned and merged to define the 3D facial template. (f) Reference 3D facial template.}
    \label{fig:depth_tracking}
\end{figure}

A subject-specific $\mathcal{X}^{\text{template}}$ was obtained using data from the face scan recording, as shown in Fig.~\ref{fig:depth_tracking}. In each data frame, we mapped $\{\mathbf{w_i}\}$ to their corresponding 3D points in the point cloud $\mathbf{x_i}\in\mathcal{X}$ via the Azure Kinect Software Development Kit (SDK)~\cite{Azure_SDK}. The convex hull of MediaPipe landmarks was then used to define the facial region of interest.
Point clouds acquired from the face scan recording were co-registered through a two-step alignment procedure. An initial coarse alignment is established based on landmark correspondences, which is subsequently refined using the Iterative Closest Point (ICP) algorithm~\cite{besl1992method}.
Following registration, $\mathcal{X}^{\text{template}}$ was obtained by averaging all aligned point clouds. 

With $\mathcal{X}$ and $\mathcal{X}^{\text{template}}$, we estimated the head pose per 
\begin{equation}
{}_{ref}T_{head} = \mbox{argmin}_{T\in SE(3)}\,d_{\text{Chamfer}}(\mathcal{X}, T\,\mathcal{X}^{\text{template}}), 
\label{eq:depth_face_registration}
\end{equation}
where $d_{\text{Chamfer}}(\mathcal{X}_1, \mathcal{X}_2)$ is the Chamfer distance between two point clouds $\mathcal{X}_1$ and $\mathcal{X}_2$, defined as
\begin{equation}
\begin{split}
d_{\text{Chamfer}}(\mathcal{X}_1, \mathcal{X}_2) 
&= \frac{1}{|\mathcal{X}_1|} \sum_{\mathbf{x_1} \in \mathcal{X}_1} \min_{\mathbf{x_2} \in \mathcal{X}_2} \|\mathbf{x_1} - \mathbf{x_2}\|_2^2 \\
&\;+\; \frac{1}{|\mathcal{X}_2|} \sum_{\mathbf{x_2} \in \mathcal{X}_2} \min_{\mathbf{x_1} \in \mathcal{X}_1} \|\mathbf{x_2} - \mathbf{x_1}\|_2^2 .
\end{split}
\end{equation}

\subsubsection{Comparison to Prior Markerless Tracking Method}\label{subsubsec:marle_method}
To benchmark the performance of the proposed methods against prior art in markerless tracking for neuronavigation, we included a comparison with the MarLe algorithm proposed by Matsuda et al.~\cite{matsuda2023marle}.
MarLe functions as a monocular RGB tracking approach. Unlike our proposed method, as in~\ref{subsubsec:rgb_mono_method}, MarLe employs the Dlib library to detect facial landmarks.
As the source code for MarLe is currently unavailable to the public, we re-implemented the method to the best of our ability.

\subsection{Personalized Head Model (PHM) Leveraging Statistical Priors}\label{subsec:personalized_head_model}\label{subsubsec:head_modeling}
All three tracking methods introduced above (Monocular RGB, Stereo RGB, and Depth) were enhanced by leveraging statistical head priors. These priors provide two key advantages in the context of TMS. First, they help regularize the template facial estimation by leveraging prior knowledge of the human head shape~\cite{ploumpis2020towards,booth20163d}. Second, they provide a complete head representation, allowing us to track any target point on the face or scalp~\cite{ploumpis2020towards, schlesinger2024scalp,schlesinger2023automatic,dai20173d}. 
\begin{figure}[t]
    \centering
    \includegraphics[width=\linewidth]{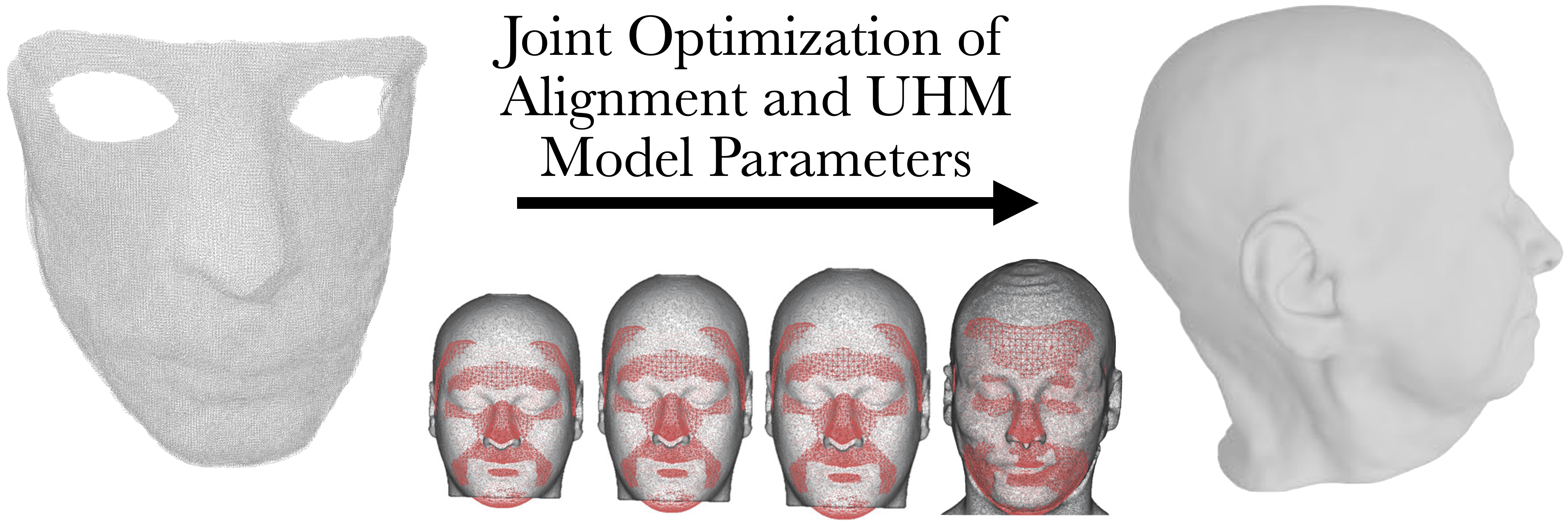}
    \caption{Personalized Head Model (PHM) created by fitting a 3D morphable model to a subject-specific facial template. The PHM is obtained by jointly aligning and optimizing model parameters. It provides a complete and annotated representation of the subject's head and face geometry.}
    \label{fig:phm}
\end{figure}

The PHM was constructed by fitting a three-dimensional morphable model (3DMM) called the Universal Head Model (UHM)~\cite{ploumpis2020towards, ploumpis2019combining} to a set of face scan point clouds. The UHM is a statistical model with a fine-detailed and complete representation of the human head structure, encompassing a broad range of subject variations. It has been annotated with the 68 Multi-PIE facial landmarks~\cite{gross2010multi} and commonly used by the Dlib library~\cite{king2009dlib}. In our previous work~\cite{schlesinger2024scalp}, we have also annotated it with the International 10--20 System neurocranial landmarks~\cite{jasper1958ten}.

The creation of the PHM for each subject involved a two-stage process. First, we created a personalized facial template as described in Section~\ref{subsubsec:depth_tracking}. Subsequently, we fitted the UHM model to this facial template by jointly optimizing a transformation and the model's coefficients, as described in Appendix~\ref{supp:subsec:3DMM_fit} and our previous work~\cite{schlesinger2024scalp}. The result is a personalized and regularized complete facial and head representation $\mathcal{X}^{PHM}$, as illustrated in Fig.~\ref{fig:phm}.

With this PHM, we implemented a second version of each of the three approaches described before (Monocular RGB, Stereo RGB, and Depth), where the PHM now defines the reference facial geometry. Monocular RGB + PHM is defined by modifying~\eqref{eq:monocular_view} to use the position of the keypoints from the PHM instead of MediaPipe. Similarly, Stereo RGB + PHM is defined by modifying~\eqref{eq:stereo_face_registration} to replace the MediaPipe keypoints $X^{\text{template}}$ with the PHM ones. Finally, Depth + PHM is defined by modifying~\eqref{eq:depth_face_registration} to use the PHM point cloud $\mathcal{X}^{PHM}$ as $\mathcal{X}^{\text{template}}$.

\subsection{Performance Characterization}\label{subsec:analysis_methods}
Performance of the proposed markerless head tracking methods was evaluated through a comparison to the head pose reported by the conventional marker-based system (NDI) throughout the video recording for each subject. Since the reference (NDI) and our tracking data were independently collected and are represented in different reference frames, pose estimates had to be synchronized and spatially aligned before comparison.

\subsubsection{Synchronization and Spatiotemporal Alignment}\label{subsubsec:synchronization_spatiotemporal_alignment} 
For both temporal and spatial alignment, similar to our previous work~\cite{green_toward_2024}, we used the trajectory of the retroreflective subject tracker as a common signal, since it could be tracked by both systems. 
Temporal alignment was achieved by maximizing the cross-correlation between the two data streams. 
For spatial alignment, since the transformation between our system and NDI's (${}_{NDI}T_{ref}$) did not vary in time (i.e., the Azure setup and the NDI system are fixed in space), we obtained it by minimizing the difference between the two trajectories after temporal alignment.
Spatial and temporal alignment parameters were then simultaneously optimized through a grid search to guarantee optimal convergence. 
Quaternions were employed throughout this process to represent rotations continuously and mitigate potential singularities associated with Euler angles.

\subsubsection{Performance Metrics}\label{subsubsec:performance_metrics}
The primary metrics we report are the root-mean-square discrepancy (RMSD) for the tracking discrepancy in the translation (mm) and angular components ($^\circ$). In addition, the robustness of each tracking system was evaluated by analyzing the failure rate, defined as the percentage of frames in which a system could not produce valid tracking data. Such failure events are attributed to various factors, including facial landmark detection errors, head pose estimation errors (e.g., when the PnP algorithm reports the head located behind the camera), and occlusions of the tracking target during extreme head orientations. 

\subsubsection{Head Pose Impact Analysis}\label{subsubsec:pose_method}
Since facial analysis tends to be sensitive to head pose~\cite{zhu2012face, ding2016comprehensive}, to evaluate the robustness of our methods against such variations, we investigated the impact of the subject's head pose on tracking performance.
Specifically, we characterized the head pose using six degrees of freedom: three translational (sway, surge, and heave) and three rotational (roll, pitch, and yaw).
For each degree of freedom, we quantified the tracking performance by computing translation and rotation RMSD independently.
This component-wise breakdown facilitates the identification of specific method shortcomings or movement types that may degrade accuracy, such as depth estimation errors during surge movements or feature occlusion during yaw rotations.

\subsection{Statistical Analysis}\label{subsec:statistical_analysis}
To assess the significance of the differences between the tracking methods, the performance metrics were analyzed with repeated measures (RM) tests reflecting the within-subject study design. Translation and rotation discrepancies served as dependent variables and were log-transformed to reduce skewness prior to statistical analysis. They were first included jointly in an RM multivariate analysis of variance (MANOVA), together with the proposed six tracking methods and the prior MarLe method~\cite{matsuda2023marle} (i.e., seven methods in total), reporting Pillai's trace. Follow-up one-way RM analysis of variance (ANOVA) tests were performed separately for translation and rotation discrepancies, where prior testing of normality and sphericity was conducted using the Shapiro-Wilk test and Mauchly's test, respectively. Upon violation of sphericity, Greenhouse–Geisser (GG) correction was employed. Due to the violation of normality in rotation discrepancy data, the non-parametric Friedman test was computed for confirmation of the ANOVA results. Following the two one-way RM ANOVA tests, post-hoc paired $t$-tests with Holm correction for multiple comparisons were applied to compare the individual tracking methods. Descriptive statistics such as the mean, median, and interquartile range were computed in log space and back-transformed to the original units.
Failure rate was reported for the six proposed methods as well as MarLe and NDI (i.e., totaling eight methods). It was analyzed using the Friedman test due to the strong skewness of the data, followed by a post-hoc pairwise Wilcoxon signed-rank test with Holm correction for multiple comparisons. Results were considered statistically significant for  $p<0.05$. Statistical analyses were performed in Python 3.10.12~\cite{van1995python} with multiple packages: RM MANOVA---R 4.5.2~\cite{r2021r} via rpy2 3.6.4~\cite{rpy2}; RM ANOVA and Mauchly’s test---Pingouin 0.5.5~\cite{vallat2018pingouin}; Holm corrections---statsmodels 0.14.5~\cite{seabold2010statsmodels}; Shapiro-Wilk, Friedman, and Wilcoxon signed-rank tests---Scipy 1.13.1~\cite{virtanen2020scipy}.

\section{Results}\label{sec:experimental_results}
\subsection{Tracking Performance}\label{subsec:tracking_results}

\begin{figure*}[t]
    \centering \includegraphics[width=0.9\linewidth]{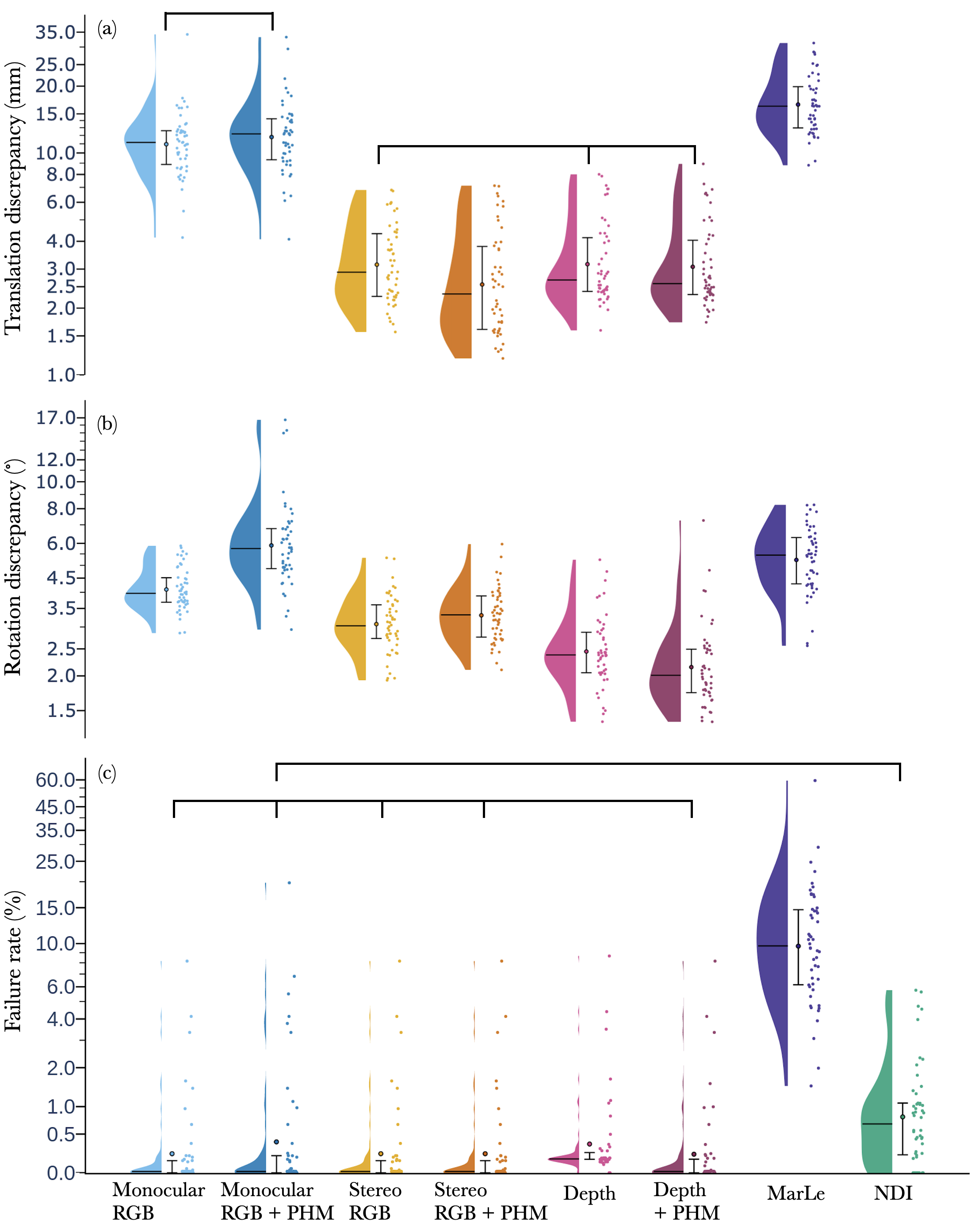}
    \caption{Performance of the head pose tracking methods evaluated empirically in 50 humans. (a) Translation and (b) rotation root-mean-square discrepancy (RMSD) of the proposed markerless methods as well as a re-implementation of a prior method, MarLe~\cite{matsuda2023marle}, relative to the conventional marker-based tracking system, NDI. (c) Failure rate is given for all methods as a percentage of the total number of frames. Violin plots show the data distribution, dots mark the mean, and box plots denote the median surrounded by the interquartile range. The y-axes are log-spaced (before transformation, $0.001\%$ was added to the data in (c) to accommodate zero values). All conditions were significantly different from each other at $p < 0.05$ (with Holm multiple comparison correction) except for the groups denoted by brackets.}
    \label{fig:tracking_results}
\end{figure*}

\begin{table*}[t]
\caption{Summary Statistics of The Performance Data of Head Pose Tracking Methods in Fig.~\ref{fig:tracking_results}}
\label{tab:statistical_analysis}
\centering
\resizebox{0.9\linewidth}{!}{
\begin{tabular}{@{}l ccc ccc cc@{}}
\toprule
\multicolumn{1}{c}{Tracking Method}
  & \multicolumn{3}{c}{Translation Discrepancy (mm)}
  & \multicolumn{3}{c}{Rotation Discrepancy ($^\circ$)} 
  & \multicolumn{2}{c}{Failure Rate (\%)} \\
  \cmidrule(lr){2-4} \cmidrule(lr){5-7} \cmidrule(lr){8-9}
  & Mean & Median & \begin{tabular}{@{}c@{}}Interquartile\\Range\end{tabular}
  & Mean & Median & \begin{tabular}{@{}c@{}}Interquartile\\Range\end{tabular}
  & Median & \begin{tabular}{@{}c@{}}Interquartile\\Range\end{tabular}\\
\midrule
Monocular RGB             
& 10.952 & 11.160 & 3.727 
& 4.094 & 3.965 & 0.829 
& \textbf{0.015}  & \textbf{0.137} \\
Monocular RGB + PHM           
& 11.760 & 12.182 & 4.950 
& 5.899 & 5.751 & 1.914 
& \textbf{0.015}  & \textbf{0.197} \\
Stereo RGB                    
&  3.141 &  2.904 & 2.073 
& 3.071 & 3.030 & 0.883 
& \textbf{0.015}  & \textbf{0.137} \\
Stereo RGB + PHM              
&  \textbf{2.555} &  \textbf{2.319} & \textbf{2.184}
& 3.304 & 3.317 & 1.122 
& \textbf{0.015}  & \textbf{0.137} \\
Depth                         
&  3.155 &  2.679 & 1.777 
& 2.448 & 2.379 & 0.817 
& 0.157  & 0.085 \\
Depth + PHM                   
&  3.071 &  2.581 & 1.741
& \textbf{2.150} & \textbf{2.011} & \textbf{0.755}
& \textbf{0.015}  & \textbf{0.153} \\
MarLe 
& 16.525 & 16.241 & 6.896 
& 5.228 & 5.446 & 2.010 
& 9.742 & 8.519 \\
NDI                           
& $-$   & $-$   & $-$  
& $-$  & $-$  & $-$  
& 0.672  & 0.866 \\
\bottomrule
\end{tabular}
}
\par\vspace{2pt}
\begin{minipage}{\dimexpr0.9\linewidth-2\tabcolsep\relax}
\footnotesize
\raggedright
Translation and rotation discrepancy statistics were calculated on the log-transformed data and back-transformed to the original units. Boldface indicates the best-performing method based on the mean for translation and rotation discrepancies and the median for failure rate.
\end{minipage}
\end{table*}

The tracking results in 50 participants demonstrated that our computer-vision-based markerless approach was capable of low discrepancy and increased robustness compared to the conventional marker-based system (NDI). The data analysis revealed a significant effect of the type of markerless tracking method on the translation and rotation discrepancy relative to NDI (MANOVA, Pillai's Trace $= 0.946$, $F_{6,44} = 129$, $p<0.0001$). There was also a significant tracking method $\times$ discrepancy metric interaction (Pillai's Trace $= 0.928$, $F_{6,44}=94.1$, $p<0.0001$), indicating the performance of each method varied differently for translation and for rotation. The subsequent one-way ANOVAs on the two discrepancy metrics confirmed a strong tracking method effect for translation discrepancy ($\varepsilon_{\mathrm{GG}}=0.417$; $F_{2.50, 123}=412$, $p<0.0001$, $\eta^2_{\mathrm{G}} = 0.774$) and rotation discrepancy ($\varepsilon_{\mathrm{GG}}=0.722$; $F_{4.33, 212}=126$, $p<0.0001$, $\eta^2_{\mathrm{G}} = 0.619$). The latter was confirmed with a Friedman test ($\chi^2_{6}=232$, $p<0.0001$; Kendall's $W=0.773$) because residual normality was violated (Shapiro–Wilk $W_{350}=0.961$, $p<0.0001$). A Friedman test also indicated that failure rate varied significantly across tracking methods ($\chi^2_{7}=222$, $p<0.0001$; Kendall's $W=0.634$). The grouped and ranked post-hoc test results are presented in Table~\ref{tab:statistical_analysis}, with detailed matrices compiled in Appendix~\ref{supp:subsec:posthoc}.

Fig.~\ref{fig:tracking_results} visualizes the experimental performance of the individual tracking methods, with statistical significance of the differences assessed by the post-hoc t-tests with multiple comparison correction. The results highlight known limitations of monocular approaches for 3D localization, specifically the lack of direct depth measurement or stereoscopic triangulation, which hinders accurate distance estimation~\cite{zhang2025survey,guo2025survey}. When the statistically-regularized personalized head models (PHM) were not employed, stereo RGB performed similarly with the depth-based approach for spatial localization ($p=1.0000$), and an inferior angular localization performance against it ($p=0.0001$).
The employment of PHM significantly improved the performance of stereo-based and depth-based tracking in different aspects, suggesting a complementary gain in tracking performance. With PHM, stereo-based tracking saw a $23\%$ decrease in the mean of translation discrepancy ($p<0.0001$), but presented an $8\%$ increase in its rotational counterpart ($p<0.0001$). Conversely, while the translation performance gain of the depth-based approach is statistically insignificant when employing PHM ($p=1.0000$), a $14\%$ decrease in the mean of rotation discrepancy was observed ($p=0.029$).
When comparing frame failure rates, stereo-RGB-camera-based and depth-sensor-based methodologies demonstrated an overall reduction in data loss compared to the NDI system (except the Monocular RGB method $p=0.096$), which, in turn, outperformed MarLe ($p<0.0001$ against all other methods). The similar failure rate of multiple proposed methods suggests the limiting factor might be landmark detection, which can fail under large head displacement angles. These results also indicate that the robustness of the depth-based tracking method when employing PHM is comparable with RGB-based methods. 
The enhanced robustness, exemplified by the lower failure rate achieved by our proposed RGB and depth-based methods, can be explained by the NDI system's reliance on the visibility of retroreflective markers, which can be occluded during natural head movements. In contrast, markerless methods leverage spatially distributed facial features with inherent redundancy, mitigating occlusion risks by eliminating the strict requirement for simultaneous visibility of all tracking points.
Furthermore, the substantial reduction in data loss relative to our re-implementation of the prior markerless method, MarLe, highlights the efficacy of the proposed algorithmic design, which demonstrates superior stability in maintaining valid pose estimation across a wider range of head orientations.

\subsection{Impact of Head Pose on Tracking}\label{subsec:components}

\begin{figure*}[htp]
        \centering
        \includegraphics[width=0.9\linewidth]{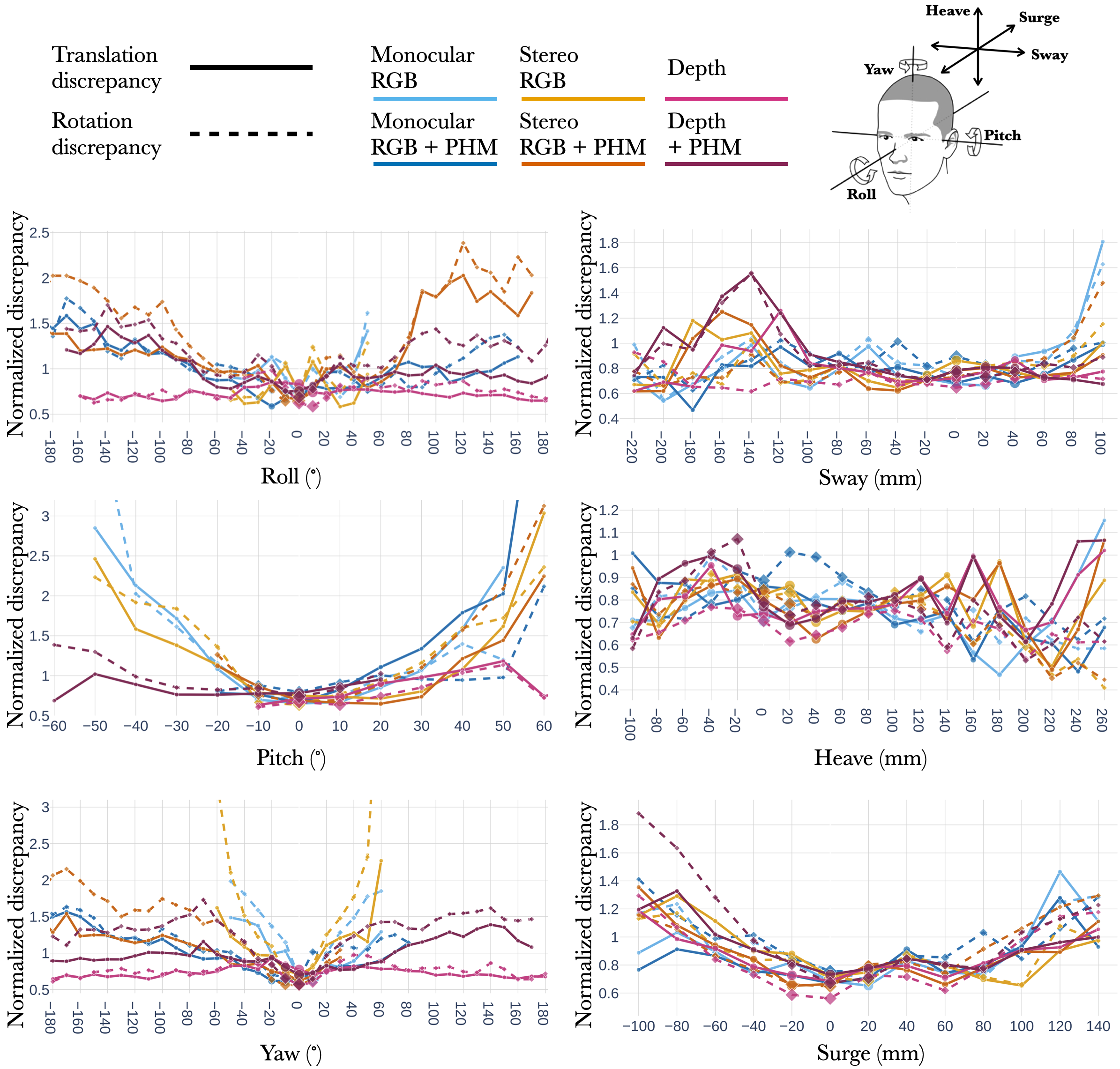}
    \caption{Mean translational and rotational tracking discrepancy as a function of the head pose parameters: sway, surge, heave, roll, pitch, and yaw. Each curve is normalized to its geometric mean value reported in Fig.~\ref{fig:tracking_results}. The size of each scatter point is scaled proportionally to the data count.}
    \label{fig:components}
\end{figure*}

Fig.~\ref{fig:components} presents the tracking performance data as a function of parameters describing the head pose.
Translational head movements introduced a slight increase in discrepancy relative to NDI across most proposed markerless tracking methods, with a generally gradual degradation.
Rotational head movements, on the other hand, exerted more pronounced effects, particularly on RGB-based tracking performance. Both Monocular and Stereo RGB methods exhibited distinct ''U-shaped'' discrepancy curves, meaning their tracking accuracy deteriorated at the extremes of rotational movements; this effect was especially evident for the pitch movements.
Conversely, depth-based methods tended to be more robust to rotational movements since they leverage a dense point cloud representation of the face.

\section{Discussion}\label{sec:discussion}

We developed methods for accurate markerless head tracking using consumer-grade RGB-D cameras. Color and depth data were combined with statistical priors to create individualized head models. Tracking performance was evaluated and compared with a conventional marker-based system in a diverse sample of 50 subjects.

Our key findings are that head pose tracking methods that incorporate depth information, whether inferred via stereo pixel correspondence or measured directly by IR depth sensors, substantially outperform Monocular RGB-based techniques in both translational and angular accuracy. In addition, they excel when subject-specific 3D priors (PHM) are employed. Interestingly, with such personalized head models based on statistical priors, we observed better translational accuracy improvement of the sparse, stereo-based method and better angular accuracy improvement of the dense, depth-based method. These findings suggest that accurate and robust markerless neuronavigation may benefit from multimodal approaches that combine the strengths of stereo RGB and depth sensors with personalized head models. 
Conversely, monocular RGB-based methods did not benefit from the use of personalized head models, likely due to the inherent limitations of monocular vision for 3D localization.

While this study utilized statistical priors fitted to camera-acquired facial geometry to construct the PHM, the proposed framework is compatible with subject-specific head models obtained via alternative modalities, including MRI, which is common in TMS research studies.
The incorporation of ground-truth anatomical data would eliminate the need for shape estimation, thereby simplifying the workflow and potentially enhancing tracking accuracy by minimizing facial geometry approximation errors.

To contextualize the observed tracking performance within clinical requirements, it is instructive to consider the operational tolerances of TMS procedures.
The dimensions of representative TMS targets are approximately 1--2 cm$^2$~\cite{koponen2018multi, giuffre2021reliability}, while cortical target localization errors typically average between 4--8 mm~\cite{lefaucheur2016value}.
Furthermore, while the NDI Polaris Vicra system used as the reference in this study has a volumetric accuracy of 0.25 mm with a 95\% confidence interval of 0.5 mm~\cite{ndi_polaris_vicra}, when accounting for registration errors, model inaccuracies, and tracker movement, conventional marker-based TMS neuronavigation systems exhibit mean errors in the range of 5.0--5.7 mm, with a 95\% confidence interval extending up to 11.5 mm~\cite{ruohonen2010navigated, Stereotaxic_instrument_2020_Neural_Navigator,nieminen2022accuracy}.
Consequently, the tracking discrepancies between our markerless methods and the conventional marker-based NDI system fall well within the error margins of the latter and the size of TMS targets, supporting the accuracy and effectiveness of our methods.

Limitations of this study include the following. (i) We relied on a relative accuracy assessment using a conventional marker-based tracking system (NDI) as a reference, which itself has inherent accuracy limitations as detailed above. (ii) We focused on comparing the accuracy and proof of concept implementation of these three complementary modalities, but did not study the computational costs or real-time performance of these methods. Future work may explore the integration of these modalities into a single unified tracking approach that leverages the strengths of each method while balancing their computational demands. (iii) The proposed methods are implemented on a per-frame basis, and adding temporal consistency constraints is likely to further improve robustness and tracking accuracy.

Beyond performance, the proposed approaches help to lower the technological and financial barriers typically associated with neuronavigation. By utilizing affordable, consumer-grade sensors and adapting established computer vision techniques, the proposed methods offer an accessible alternative to conventional stereotactic systems, which require costly cameras and rigid tracker attachment. This affordability makes the system suitable for deployment in resource-limited settings and large-scale research studies, where conventional neuronavigation may be impractical.

\section{Conclusions}\label{sec:conclusions}
Markerless head pose tracking using consumer-grade RGB-D sensor technology is feasible and can achieve localization accuracy suitable for TMS neuronavigation. Tracking methods that incorporate depth information and subject-specific head models significantly outperform monocular techniques in accuracy. These novel methods also improve upon the tracking failure rate compared to both markerless monocular approaches as well as the widely used marker-based stereotaxy solutions, especially during extreme head pose deviations. There are opportunities to combine stereo RGB and depth sensor information with personalized statistical head models to enhance markerless tracking further. By eliminating the need for specialized camera hardware, physical marker attachment, and manual registration, the proposed markerless approach can substantially reduce setup complexity and cost. Overall, this work advances a new class of accessible, accurate, and non-invasive head tracking systems that leverage affordable, consumer-grade sensors and advanced modeling algorithms.

\appendices

\appendix

\subsection{Experimental Setup}\label{supp:subsec:experimental_setup}
\subsubsection{Acquisition Hardware}\label{supp:subsubsec:data_acquisition_hardware}
Reference data were obtained using the NDI Polaris Vicra optical measurement system. This stereotaxy camera system utilizes IR light to detect a set of retroreflective markers attached to the subject's head. Experimental data were captured using two Microsoft Azure Kinect DK devices running in Master and Subordinate mode. Each Azure device records both RGB and infrared/depth data simultaneously, at 30 frames per second (FPS) in both domains and with respective resolutions of $3840\times2160$ and $640\times576$ pixels.

Both Azure Kinect DK devices were mounted on a 3.18 mm-thick aluminum L-channel, oriented coplanarly but offset horizontally by 355 mm. The NDI Polaris Vicra was also mounted to this system to ensure all collected 3D data were rigidly related. All data collection devices were connected to USB 3.0+ controllers on a Dell XPS 8950 desktop computer (i7-12700 @ 2.1-4.9 GHz, 32 GB RAM, RTX 3090 GPU).

\subsubsection{Acquisition Software}\label{supp:subsubsec:data_acquisition_software}
Real-time dataset acquisition was achieved using a custom multi-threaded C++ application. The Microsoft Azure Kinect SDK was used to interface with the Azure Kinect DK devices. The NDI Combined API C Interface Library~\cite{NDI_Combined_API_C_Interface} was used to interface with the NDI Polaris Vicra. The libjpeg-turbo library~\cite{libjpeg-turbo} was used to decode RGB data, which was natively transmitted in MJPEG format by the Azure Kinect DK devices. Data captured by the Azure Kinect DK devices was written to disk by the Azure Kinect SDK in Matroska Multimedia Container format. Data captured by the NDI Polaris Vicra was reported as homogeneous $4\times4$ transformation matrices (1 matrix per tracker per time point) and written to disk in plain text format.

\subsubsection{Camera Calibration}\label{supp:subsubsec:camera_calibration}
Camera calibration was achieved using a custom, OpenCV-based~\cite{opencv_library} pipeline as follows. A $9\times13$ chessboard pattern (square edge is 25.4 mm long) was printed on a 5 mm thick foam board. Intrinsic and extrinsic calibration data for both cameras were then acquired simultaneously by moving the chessboard poster in various positions and orientations throughout the fields of view of both cameras. Both Azure Kinect DK devices were configured to record RGB data synchronously at $3840\times2160$ resolution and 30 FPS with an exposure time of 2.5 ms per frame. IR data was captured in passive IR mode at $1024\times1024$ resolution and 30 FPS with a 1.6 ms exposure time. Ambient illumination was increased with artificial lighting to maintain proper image exposure. Inter-device synchronization was achieved using the manufacturer's firmware via a wired connection. 

First, intrinsic parameters were estimated for each camera using only those video frames for which the chessboard was in view of the individual camera (OpenCV findChessboardCorners function~\cite{opencv_library}). Second, stereo calibration was performed (OpenCV stereoCalibrate function) using only those pairs of frames for which the chessboard was detected in both cameras. Frame pairs were synchronized across Azure Kinect DK devices using hardware-based synchronization data. As the original recordings were at 30 FPS (i.e., consecutive frames have high correlation), we split the data into 10 datasets equivalent to a 3 FPS recording (taking 1 every ten frames). Effectively, 10 intrinsic calibrations were performed for each camera, and the final camera matrix and distortion coefficients were obtained by element-wise averaging the output of the 10 trials. Similarly, after synchronization, the stereo calibration dataset was split into 10 smaller datasets, and the calibration matrices were obtained by element-wise averaging of the output of the 10 trials.

\subsection{Facial Landmark Selection for RGB-based Tracking Methods}\label{supp:subsec:rgb_ablation_study}

\begin{figure}[t]
  \centering
\resizebox{\linewidth}{!}{%
\begin{minipage}{\linewidth}
  \begin{subfigure}[b]{0.24\textwidth}
    \centering
    \includegraphics[width=\linewidth]{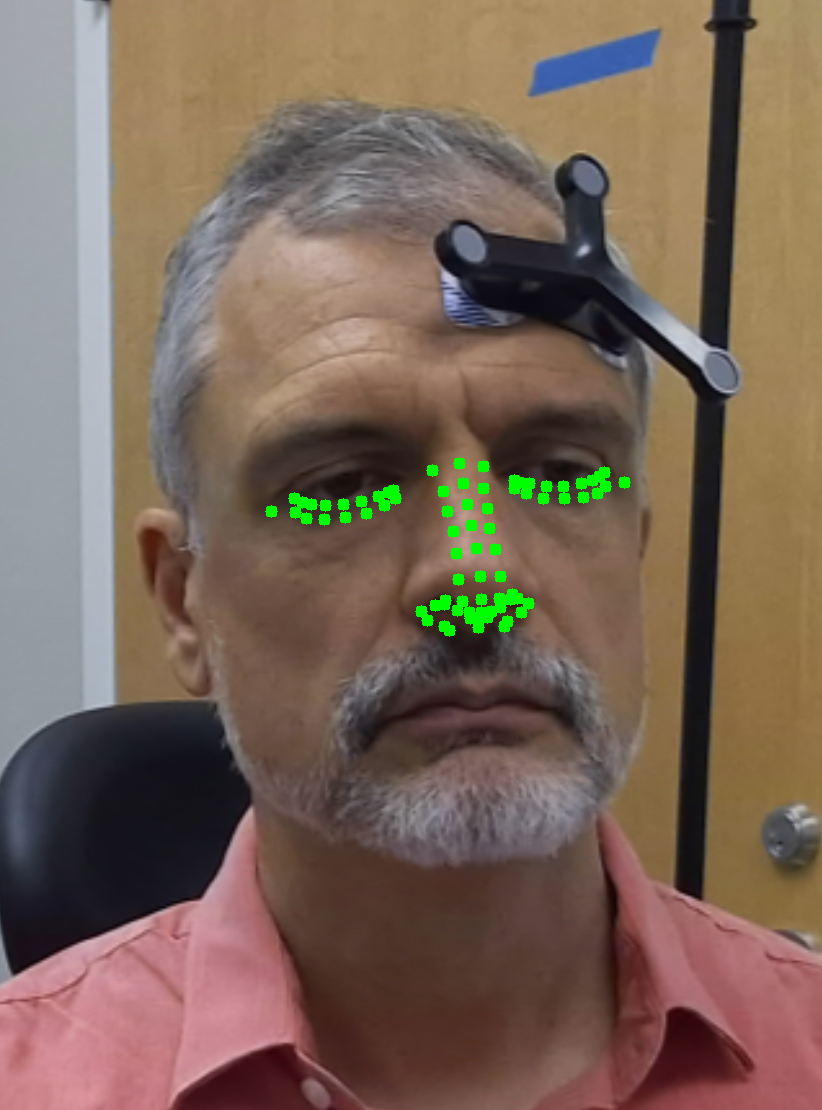}
    \caption{}\label{fig:landmark_group1}
  \end{subfigure}\hfill
  \begin{subfigure}[b]{0.24\textwidth}
    \centering
    \includegraphics[width=\linewidth]{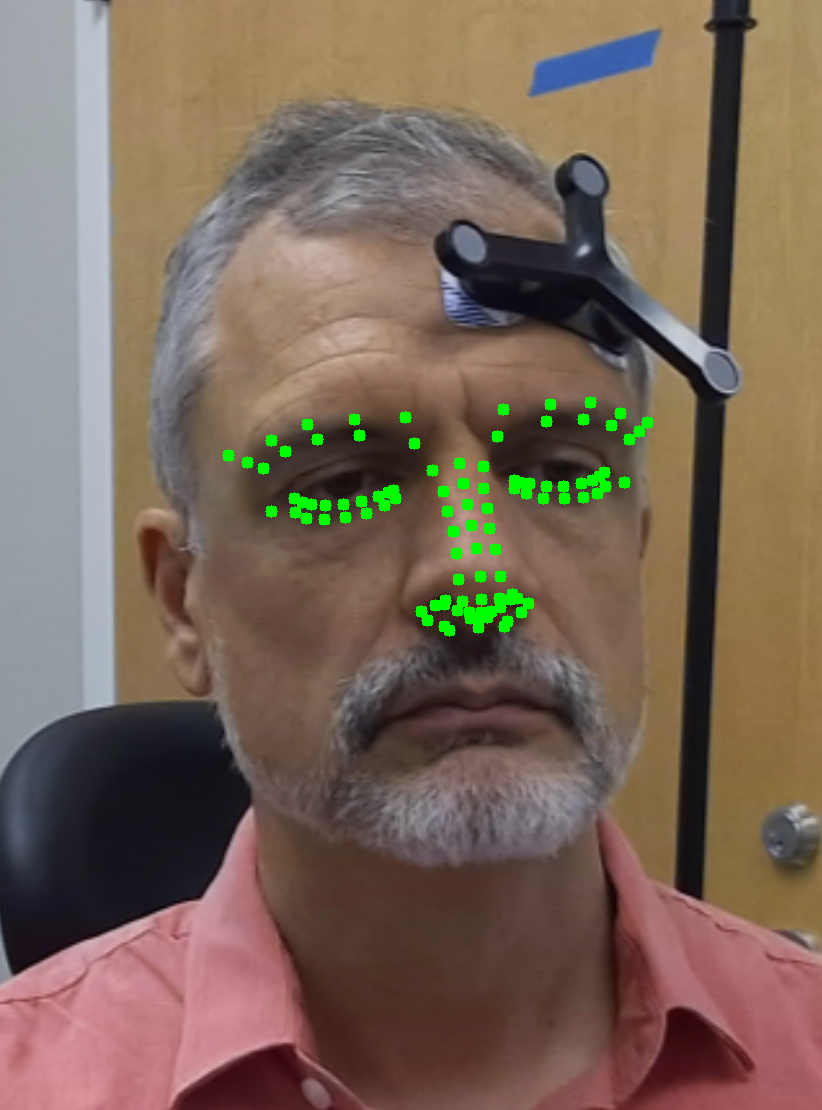}
    \caption{}\label{fig:landmark_group2}
  \end{subfigure}\hfill
  \begin{subfigure}[b]{0.24\textwidth}
    \centering
    \includegraphics[width=\linewidth]{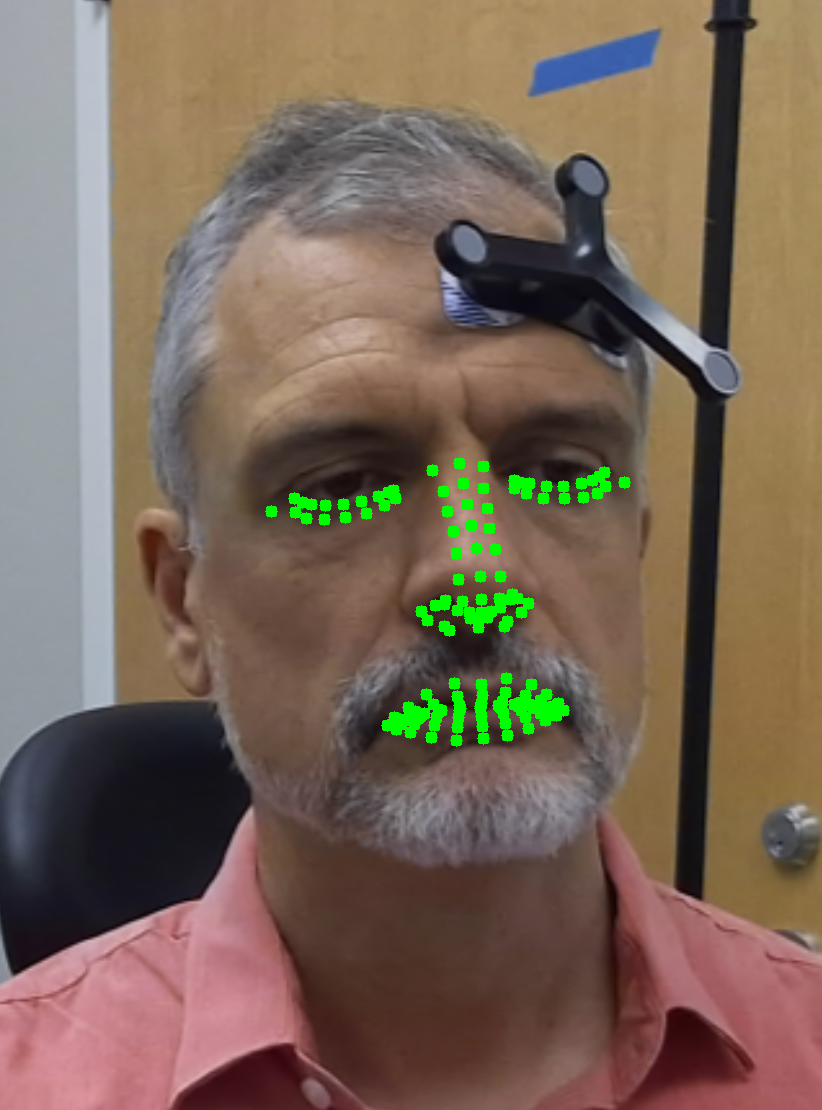}
    \caption{}\label{fig:landmark_group3}
  \end{subfigure}\hfill
  \begin{subfigure}[b]{0.24\textwidth}
    \centering
    \includegraphics[width=\linewidth]{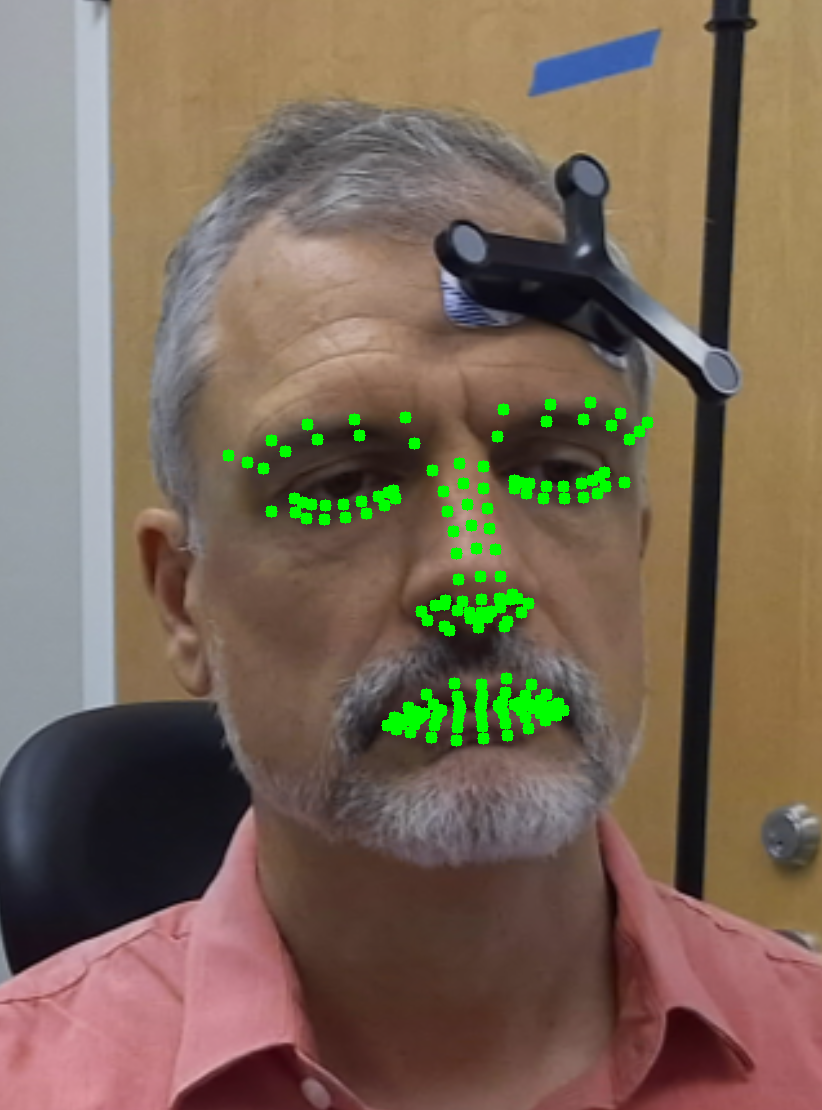}
    \caption{}\label{fig:landmark_group4}
  \end{subfigure}
  \\[1mm]
  \begin{subfigure}[b]{0.24\textwidth}
    \centering
    \includegraphics[width=\linewidth]{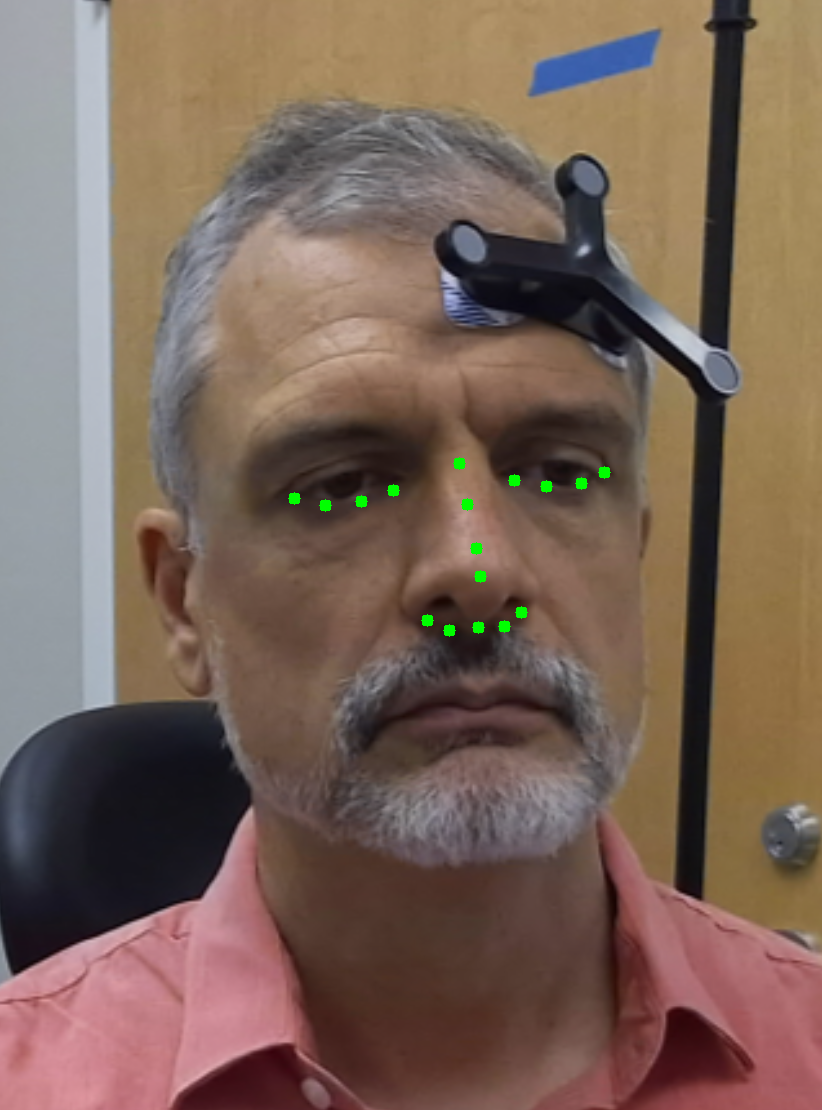}
    \caption{}\label{fig:landmark_group5}
  \end{subfigure}\hfill
  \begin{subfigure}[b]{0.24\textwidth}
    \centering
    \includegraphics[width=\linewidth]{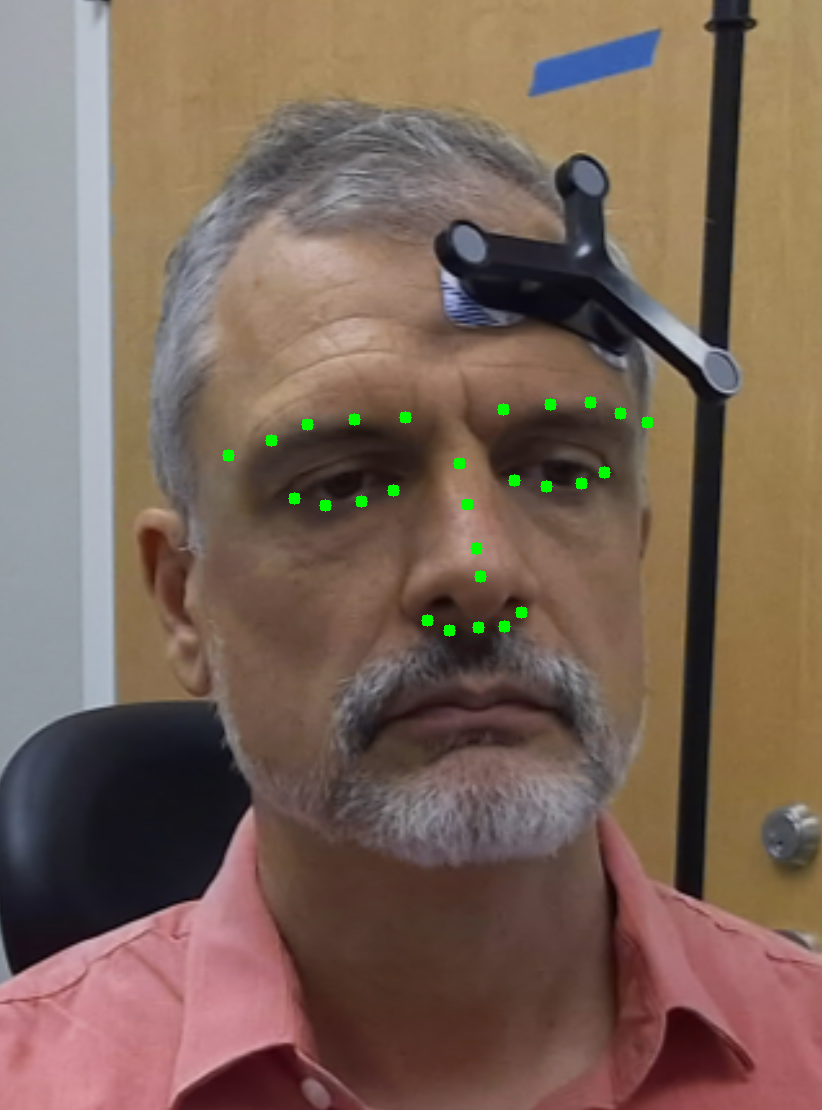}
    \caption{}\label{fig:landmark_group6}
  \end{subfigure}\hfill
  \begin{subfigure}[b]{0.24\textwidth}
    \centering
    \includegraphics[width=\linewidth]{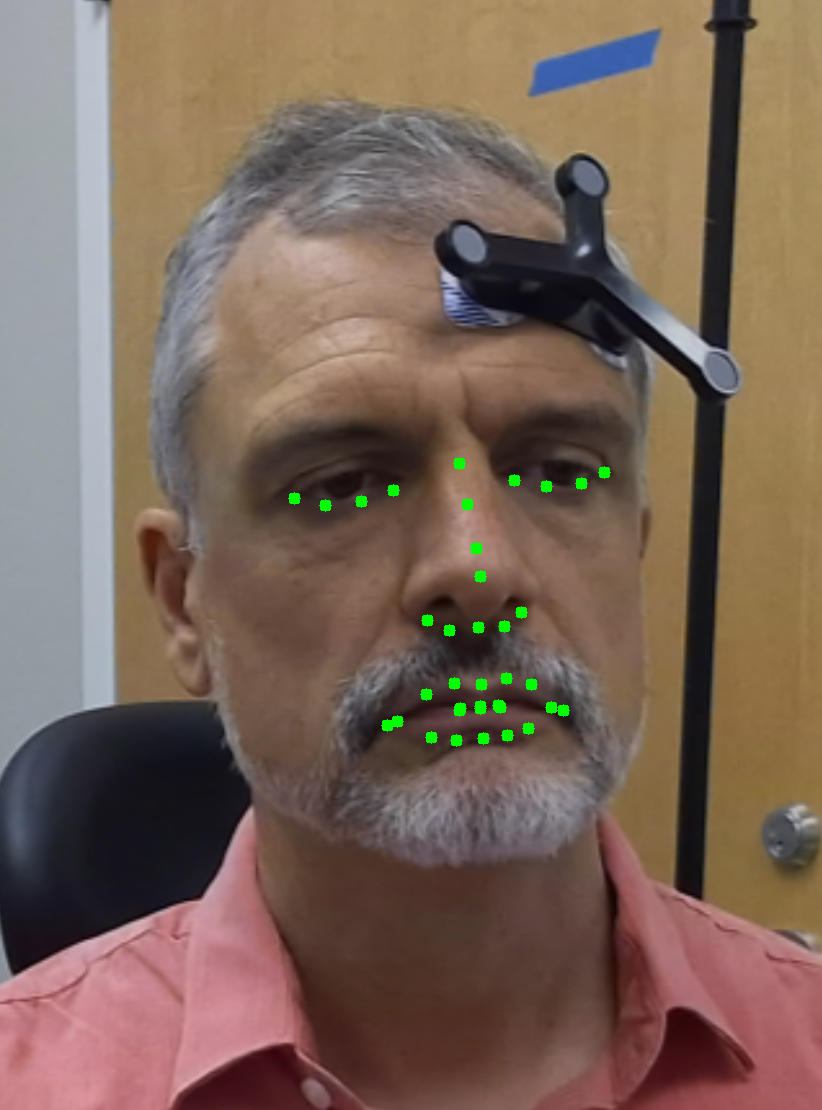}
    \caption{}\label{fig:landmark_group7}
  \end{subfigure}\hfill
  \begin{subfigure}[b]{0.24\textwidth}
    \centering
    \includegraphics[width=\linewidth]{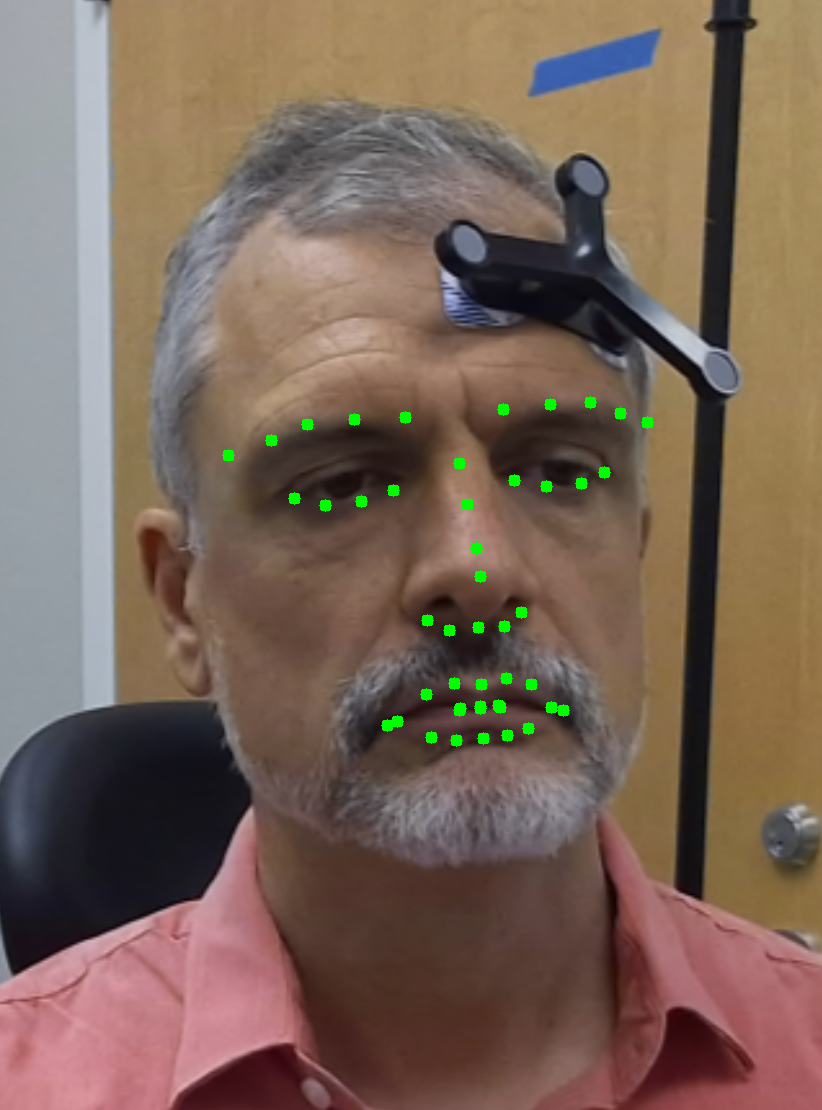}
    \caption{}\label{fig:landmark_group8}
  \end{subfigure}
  \end{minipage}%
  }

  \caption{Visualization of the eight facial landmark combinations with dense landmarks on top and the Multi-PIE compatible sparse versions on the bottom. (a) Selected lower eye and nose area landmarks. (b) Selected lower eye, nose and eyebrow area landmarks. (c) Selected lower eye, nose and mouth area landmarks. (d) A union set of (a), (b) and (c). (e) Multi-PIE-compatible sparse version of (a). (f) Multi-PIE-compatible sparse version of (b). (g) Multi-PIE-compatible sparse version of (c). (h) A union set of (e), (f) and (g).}
  \label{fig:landmarks_rgb_supplement}
\end{figure}
The use of RGB-based methods in Section~\ref{subsubsec:rgb_mono_method} and Section~\ref{subsubsec:rgb_stereo_method} relies on facial landmarks. Hence, it is critical to determine which facial landmark subsets could be used for accurate subject head tracking. Salient facial areas, including eyebrows, eyes, nose, and mouth, were empirically observed to have a higher detection accuracy over significant head movements. We used facial landmarks that cover the lower part of the eyes and the nose as the base subset of the facial landmarks, and compared the tracking accuracy of the RGB-based methods between them and when additional facial landmarks, such as those covering the nose or the mouth, are involved. Specifically, four subsets of the total detected facial landmarks were selected, covering the following areas: the lower part of the eyes and the nose; the lower part of the eyes, the nose and the eyebrows; the lower part of the eyes, the nose and the mouth; and the lower part of the eyes, the nose, the eyebrows and the mouth. When the PHM was used as the 3D head model, a sparse subset of these detected landmarks that is Multi-PIE compatible was used. A visualization of the set of landmarks tested is shown in Fig.~\ref{fig:landmarks_rgb_supplement}, with corresponding tracking performance statistics provided in Table~\ref{tab:rgb_landmarks}.

\begin{table*}[!htbp]
\caption{RGB Tracking across Landmark Subsets}
\label{tab:rgb_landmarks}
\centering
\resizebox{0.9\linewidth}{!}{
\begin{tabular}{@{}c@{\hspace{1em}}lcccccc@{}}
\toprule
& \multicolumn{1}{c}{Landmark Subsets}
& \multicolumn{3}{c}{\makecell{Monocular RGB\\Euclidean Distance RMSD (mm)}}
& \multicolumn{3}{c}{\makecell{Stereo RGB\\Euclidean Distance RMSD (mm)}} \\
\cmidrule(lr){3-5}\cmidrule(lr){6-8}
& \multicolumn{1}{c}{}
& Median & Mean & SD
& Median & Mean & SD \\
\midrule
\multirow{8}{*}{\rotatebox{90}{}} 
& Eyes + Nose (Fig.~\ref{fig:landmarks_rgb_supplement}(\subref{fig:landmark_group1})) & 12.345 & 12.838 & 4.012 & 3.581 & 4.052 & 1.318\\
& Eyes + Nose + Eyebrows (Fig.~\ref{fig:landmarks_rgb_supplement}(\subref{fig:landmark_group2})) & 11.707 & 12.045 & 3.559 & 3.454 & 3.947 & 1.389\\
& Eyes + Nose + Mouth (Fig.~\ref{fig:landmarks_rgb_supplement}(\subref{fig:landmark_group3})) & 12.380 & 12.897 & 4.831 & 3.022 & 3.491 & 1.414\\
& Eyes + Nose + Eyebrows + Mouth (Fig.~\ref{fig:landmarks_rgb_supplement}(\subref{fig:landmark_group4})) & 11.160 & 11.582 & 4.403 & 2.904 & 3.410 & 1.444\\
& Eyes + Nose $^{\mathrm{a}}$ (Fig.~\ref{fig:landmarks_rgb_supplement}(\subref{fig:landmark_group5})) & 12.580 & 13.011 & 4.024 & 3.454 & 3.956 & 1.330\\
& Eyes + Nose + Eyebrows $^{\mathrm{a}}$  (Fig.~\ref{fig:landmarks_rgb_supplement}(\subref{fig:landmark_group6})) & 11.572 & 11.916 & 3.496 & 3.411 & 3.881 & 1.405\\
& Eyes + Nose + Mouth $^{\mathrm{a}}$ (Fig.~\ref{fig:landmarks_rgb_supplement}(\subref{fig:landmark_group7})) & 12.788 & 13.373 & 5.080 & 2.989 & 3.503 & 1.415\\
& Eyes + Nose + Eyebrows + Mouth $^{\mathrm{a}}$ (Fig.~\ref{fig:landmarks_rgb_supplement}(\subref{fig:landmark_group8})) & 11.118 & 11.689 & 4.164 & 2.894 & 3.444 & 1.446\\
\midrule
\multirow{4}{*}{\rotatebox{90}{PHM}} 
& Eyes + Nose $^{\mathrm{a}}$ (Fig.~\ref{fig:landmarks_rgb_supplement}(\subref{fig:landmark_group5})) & 18.922 & 19.065 & 5.107 & 2.417 & 3.089 & 1.699\\
& Eyes + Nose + Eyebrows $^{\mathrm{a}}$ (Fig.~\ref{fig:landmarks_rgb_supplement}(\subref{fig:landmark_group6})) & 14.324 & 14.675 & 4.652 & 2.438 & 3.078 & 1.712\\
& Eyes + Nose + Mouth $^{\mathrm{a}}$ (Fig.~\ref{fig:landmarks_rgb_supplement}(\subref{fig:landmark_group7})) & 15.448 & 15.707 & 5.234 & 2.421 & 3.049 & 1.706\\
& Eyes + Nose + Eyebrows + Mouth $^{\mathrm{a}}$ (Fig.~\ref{fig:landmarks_rgb_supplement}(\subref{fig:landmark_group8})) & 12.182 & 12.667 & 5.258 & 2.325 & 2.955 & 1.722\\
\bottomrule
\end{tabular}
}
\par\vspace{2pt}
\begin{minipage}{\dimexpr0.9\linewidth-2\tabcolsep\relax}
\footnotesize
\raggedright
Unlike the statistics in Table~\ref{tab:statistical_analysis}, raw values are reported here.\\
$^{\mathrm{a}}$Sparse subset of landmarks approximating the Dlib landmark configuration.
\end{minipage}
\end{table*}

\subsection{Subject Representations}\label{supp:subsec:3DMM_fit}
\subsubsection{Three-Dimensional Morphable Models}\label{supp:subsubsec:3DMM}
Three-dimensional morphable models (3DMMs) are statistical models that capture object class variability. They are typically constructed by acquiring and aligning numerous 3D object scans.
A 3DMM is characterized by its mean shape, $S_{mean}$, and a set of $K$ principal component vectors, denoted $\mathrm{v} = [v_1\ ...\ v_k]$, $v_i \in \mathbb{R}^n$, and their corresponding eigenvalues, $\mathrm{e} = [e_1\ ...\ e_k]$, obtained through principal component analysis (PCA).
These components represent dominant shape variation modes, maintaining dense correspondence with $S_{mean}$.
New object instances, $S_w$, can be generated by linearly combining the mean shape with a weighted sum, $\mathrm{w}$, of the eigenvalues multiplied by eigenvectors with customized weights:
\begin{equation}
    \mathrm{w} = [w_1\ ...\ w_k], w_i \in \mathbb{R},
\end{equation}
\begin{equation}
    S_w = S_{mean}+\sum_{i=1}^{k}{w_i \cdot e_i \cdot v_i}.
\end{equation}
This structure allows for a comprehensive yet efficient representation of object instances by retaining only the most significant variations, thereby substantially reducing the parameter space.

\subsubsection{3DMM Fitting Procedure}\label{supp:subsubsec:fitting_details}
The modeling procedure begins with the selection of a subset of consecutive frames from the depth sensor face scan recording, followed by the detection of facial landmarks within these frames. These frames are not part of any future tracking steps. Let $\mathcal{N} = {n_1, n_2, \dots, n_N}$ represent the set of selected frames. For each frame $n_k \in \mathcal{N}$, the detected 2D facial landmarks from the left Azure RGB images are denoted as $\mathbf{p}^L[n] = {p^L_l [n]}^{468}_{l=1}$, as described in Section~\ref{subsubsec:face_alignment}. Using the extracted convex hull from $\mathbf{p}^L[n_k]$, the corresponding depth data is masked to obtain the dense facial point clouds. Let $\mathbf{D}_{face}^L[n_k]$ represent the segmented 3D facial point clouds from the left Azure device at frame $n_k$. To capture the subject-specific 3D facial geometry, the full set of selected point clouds is aggregated:
\begin{equation}
    \mathbf{D}_{face} = \bigcup_{\mathcal{N}} \left( \mathbf{D}_{face}^L[n_k]\right).
\end{equation}
These point clouds are subsequently aligned using a multi-way registration method~\cite{choi2015robust} to produce an aggregated and downsampled point cloud, denoted as $\mathcal{X}_{face}$. This point cloud serves as a representation of the subject’s facial geometry captured by the Azure system.

Subsequently, an optimization framework inspired by Schlesinger et al.~\cite{schlesinger2024scalp} was employed to deform the 3DMM template to fit the aggregated point cloud, $\mathcal{X}_{face}$, derived from Azure-captured point clouds as described in Section~\ref{subsubsec:depth_tracking}.
This process involves computing a set of weights to fit the 3DMM to the subject face geometry, $w$, and computing the transformation that aligns both point clouds, $T$. This optimization, capturing the unique facial geometry and yielding the subject-specific model, can be formulated as the minimization of the objective function provided in:
\begin{equation}
    l(w, T) = d_{\text{Chamfer}}(\mathcal{X}_{face}, \mathcal{X}_{UHM}) + \lambda \left( {\lVert w \rVert_F} + {\lVert T-T_0 \rVert_F} \right),
    \label{eq:loss_3dmm}
\end{equation}
 where $d_{\text{Chamfer}}$ quantifies the geometric discrepancy between two point clouds by measuring the average distances between corresponding points and the angles between their respective surface normals, as defined by the Chamfer distance metric~\cite{barrow1977parametric}; the second term acts as a regularization factor penalizing high deformations or transformations. 

\subsubsection{Rigid Alignment of Face and Head Models}\label{supp:subsubsec:mappings}
Since the 3D facial and head models used in this study have different coordinate systems, e.g., origins for each PHM are at the nasion, and their counterpart for the MediaPipe Face Model (MP) is the center of their internal head model~\cite{kartynnik_real-time_2019}, it is necessary to ensure that all tracking methods report transformations towards the same coordinate frame. We selected the coordinate frame of the PHM as the common coordinate frame of the head, and estimated the rigid transformation to map to this reference frame from other models when needed.

\subsection{Statistical Analysis Post-Hoc Pairwise Comparison P-Values}\label{supp:subsec:posthoc}
P-values of the post-hoc pairwise comparisons are reported in Tables~\ref{tab:posthoc_translation},~\ref{tab:posthoc_rotation} and~\ref{tab:posthoc_failure_rate}.

\begin{table*}[t]
\caption{Post-hoc Pairwise Comparisons for Translation Discrepancy ($p$-values)}
\label{tab:posthoc_translation}
\centering
\resizebox{0.9\linewidth}{!}{
\begin{tabular}{@{}l ccc ccc ccc@{}}
\toprule
\multicolumn{1}{c}{Tracking Method}
  & \multicolumn{1}{c}{Monocular}
  & \multicolumn{1}{c}{Monocular} 
  & \multicolumn{1}{c}{Stereo}
  & \multicolumn{1}{c}{Stereo}
  & \multicolumn{1}{c}{Depth}
  & \multicolumn{1}{c}{Depth}
  & \multicolumn{1}{c}{MarLe}
  \\
  & {RGB}
  & {RGB + PHM} 
  & {RGB}
  & {RGB + PHM}
  & {}
  & {+ PHM}
  & {}
  \\
\midrule
Monocular RGB             
& $-$ & {} & {}& {} 
& {} & {} & {} \\
Monocular RGB + PHM           
& 0.0801 & $-$ &  &  
&  &  &  \\
Stereo RGB                    
&  $<$0.0001 &  $<$0.0001 & $-$ &  
&  &  &  \\
Stereo RGB + PHM              
&  $<$0.0001 &  $<$0.0001 & $<$0.0001 & $-$ 
&  &  &  \\
Depth                         
&  $<$0.0001 &  $<$0.0001 & 1.0000 & $<$0.0001 
& $-$ &  & \\
Depth + PHM                   
&  $<$0.0001 &  $<$0.0001 & 1.0000 & 0.0004 
& 1.0000 & $-$ & \\
MarLe 
&  $<$0.0001 &  $<$0.0001 & $<$0.0001 & $<$0.0001 
& $<$0.0001 & $<$0.0001 & $-$ \\
\bottomrule
\end{tabular}
}
\par\vspace{2pt}
\begin{minipage}{\dimexpr0.9\linewidth-2\tabcolsep\relax}
\footnotesize
\raggedright
Values shown correspond to tracking method rankings in Table~\ref{tab:statistical_analysis}.
\end{minipage}
\end{table*}

\begin{table*}[t]
\caption{Post-hoc Pairwise Comparisons for Rotation Discrepancy ($p$-values)}
\label{tab:posthoc_rotation}
\centering
\resizebox{0.9\linewidth}{!}{
\begin{tabular}{@{}l ccc ccc ccc@{}}
\toprule
\multicolumn{1}{c}{Tracking Method}
  & \multicolumn{1}{c}{Monocular}
  & \multicolumn{1}{c}{Monocular} 
  & \multicolumn{1}{c}{Stereo}
  & \multicolumn{1}{c}{Stereo}
  & \multicolumn{1}{c}{Depth}
  & \multicolumn{1}{c}{Depth}
  & \multicolumn{1}{c}{MarLe}
  \\
  & {RGB}
  & {RGB + PHM} 
  & {RGB}
  & {RGB + PHM}
  & {}
  & {+ PHM}
  & {}
  \\
\midrule
Monocular RGB             
& $-$ &  &  &  
&  &  &  \\
Monocular RGB + PHM           
& $<$0.0001 & $-$ &  & 
& &  & \\
Stereo RGB                    
&  $<$0.0001 &  $<$0.0001 & $-$ &  
& &  &  \\
Stereo RGB + PHM              
&  $<$0.0001 &  $<$0.0001 & $<$0.0001 & $-$ 
&  &  &  \\
Depth                         
&  $<$0.0001 &  $<$0.0001 & 0.0001 & $<$0.0001 
& $-$ &  &  \\
Depth + PHM                   
&  $<$0.0001 &  $<$0.0001 & $<$0.0001 & $<$0.0001 
& 0.0289 & $-$ &  \\
MarLe 
&  $<$0.0001 &  0.0326 & $<$0.0001 & $<$0.0001 
& $<$0.0001 & $<$0.0001 & $-$ \\
\bottomrule
\end{tabular}
}
\par\vspace{2pt}
\begin{minipage}{\dimexpr0.9\linewidth-2\tabcolsep\relax}
\footnotesize
\raggedright
Values shown correspond to tracking method rankings in Table~\ref{tab:statistical_analysis}.
\end{minipage}
\end{table*}

\begin{table*}[t]
\caption{Post-hoc Pairwise Comparisons for Failure Rate ($p$-values)}
\label{tab:posthoc_failure_rate}
\centering
\resizebox{0.9\linewidth}{!}{
\begin{tabular}{@{}l ccc ccc ccc@{}}
\toprule
\multicolumn{1}{c}{Tracking Method}
  & \multicolumn{1}{c}{Monocular}
  & \multicolumn{1}{c}{Monocular} 
  & \multicolumn{1}{c}{Stereo}
  & \multicolumn{1}{c}{Stereo}
  & \multicolumn{1}{c}{Depth}
  & \multicolumn{1}{c}{Depth}
  & \multicolumn{1}{c}{MarLe}
  & \multicolumn{1}{c}{NDI}
  \\
  & {RGB}
  & {RGB + PHM} 
  & {RGB}
  & {RGB + PHM}
  & {}
  & {+ PHM}
  & {}
  & {}
  \\
\midrule
Monocular RGB             
& $-$ &  &  &  
&  &  &  \\
Monocular RGB + PHM           
& 0.0840 & $-$ &  & 
& &  & \\
Stereo RGB                    
&  1.0000 &  0.0840 & $-$ &  
& &  &  \\
Stereo RGB + PHM              
&  1.0000 &  0.0840 & 1.0000 & $-$ 
&  &  &  \\
Depth                         
&  $<$0.0001 &  0.0077 & $<$0.0001 & $<$0.0001 
& $-$ &  &  \\
Depth + PHM                   
&  1.0000 &  0.1911 & 1.0000 & 1.0000 
& $<$0.0001 & $-$ &  \\
MarLe 
&  $<$0.0001 &  $<$0.0001 & $<$0.0001 & $<$0.0001 
& $<$0.0001 & $<$0.0001 & $-$ \\
NDI 
&  0.0023 &  0.0961 & 0.0023 & 0.0023
& 0.0046 & 0.0012 & $<$0.0001 & $-$ \\
\bottomrule
\end{tabular}
}
\par\vspace{2pt}
\begin{minipage}{\dimexpr0.9\linewidth-2\tabcolsep\relax}
\footnotesize
\raggedright
Values shown correspond to tracking method rankings in Table~\ref{tab:statistical_analysis}.
\end{minipage}
\end{table*}

\section*{Acknowledgment}
For contributions to earlier stages of this project, we want to acknowledge Rohit Raguram, Ravitashaw Bathla, Zhuoqing Chang, Alexander Shang, Raymond Lin, and Dmitry Isaev. We also thank Dr.~Joan Camprodon for advice on relevant clinical developments.

\section*{References}


\end{document}